# Embodied intelligent industrial robotics: Framework and techniques*

**Chaoran Zhang[a,†], Chenhao Zhang[a,†], Zhaobo Xu[a], Qinghongbing Xie[a], Jinliang Hou[b], Pingfa Feng[a,c], Long Zeng[a,*]**

a Tsinghua Shenzhen International Graduate School, Tsinghua University, Shenzhen 518055, China

b Guangzhou Fuwei Intelligent Technology Co., Ltd, Guangzhou 510760, China

c Department of Mechanical Engineering, Tsinghua University, Beijing 100084, China

* Corresponding author. E-mail address: zenglong@sz.tsinghua.edu.cn

† These authors equally contributed to this work.

**Abstract:**

The combination of embodied intelligence and robots has great prospects and is becoming increasingly common. In order to work more efficiently, accurately, reliably, and safely in industrial scenarios, robots should have at least general knowledge, working-environment knowledge, and operating-object knowledge. These pose significant challenges to existing embodied intelligent robotics (EIR) techniques. Thus, this paper first briefly reviews the history of industrial robotics and analyzes the limitations of mainstream EIR frameworks. Then, a new knowledge-driven technical framework of embodied intelligent industrial robotics (EIIR) is proposed for various industrial environments. It has five modules: a world model, a high-level task planner, a low-level skill controller, a simulator, and a physical system. The world model provides industrial working-environment knowledge (such as semantic maps) and industrial operating-object knowledge (such as knowledge graphs) that existing multi-modal large language models lack. The high-level task planner decomposes tasks described with natural language into a series of subtasks. The low-level skill controller translates these subtasks into specific physical-executable sequential skills. The simulator models the kinematics, control logic, and environmental interactions, thereby enabling algorithm development, virtual commissioning, and digital twins at both single-robot and full-production-line scales. The development of techniques related to each module are also thoroughly reviewed, and recent progress regarding their adaption to industrial applications are discussed. A case study of real-world assembly system is given to demonstrate the newly proposed EIIR framework's applicability and potentiality. Finally, the key challenges that EIIR encounters in industrial scenarios are summarized and future research directions are suggested. The authors believe that EIIR technology is shaping the next generation of industrial robotics and EIIR-based industrial systems supply a new technological paradigm for intelligent manufacturing. It is expected that this review could serve as a valuable reference for scholars and engineers that are interested in industrial embodied intelligence. Together, scholars can use this research to drive their rapid advancement and application of EIIR techniques. The authors would continue to track and contribute new studies in the project page https://github.com/jackyzengl/EIIR.

# 1 Introduction

With the explosive development of embodied intelligence (EI, also known as embodied artificial intelligence) [1, 2], its combination with robotics—collectively referred to as embodied intelligent robotics (EIR) [3-5]—has stronger humanoid abilities and can perform more general tasks, i.e. task-level flexibility. Most current EIR studies focus on the application to daily-life scenarios, such as home service and social interaction [6]. However, industrial scenarios usually require techniques to have much higher efficiency, accuracy, reliability, and safety, and these challenges have been recognized by several academic studies [5, 7, 8]. Thus, we aim to systematically study the potential combination of EI and industrial robotics (IR) techniques, i.e. embodied intelligent industrial robotics (EIIR[1]). EIIR primarily focuses on industrial agents that can independently perceive, make decisions, and execute tasks within industrial environments. The further development of EIIR has great potential to become a new technological branch and a new paradigm of innovative intelligent manufacturing technology.

The most fundamental reason for the challenges faced by existing EIR when transferred directly to industrial scenarios, is the lack of industrial knowledge for existing multi-modal large language models (MLLMs) [9-12]. For an industrial robot to work efficiently, accurately, and reliably, we argue that it must possess at least three kinds of knowledge, i.e. general knowledge, working-environment knowledge, and operating-object knowledge. The general knowledge enables robots to communicate naturally and reasonably with humans. The working-environment knowledge makes robots to know its working environment well and locomote efficiently. The operating-object knowledge supplies rich domain-specific knowledge for well manipulation, satisfying industrial standards and constraints. Thus, based on our insights, we further propose a new knowledge-driven EIIR technical framework and give a thoroughly review accordingly. We expected our work could serve as a valuable review for scholars and engineers who are interested in this prospective area.

According to a literature search in the Scopus database that utilized the "embodied intelligence" and "embodied intelligence AND (manufacturing OR industrial)" keywords, two clear trends in the research have emerged over the past half-century. **Fig. 1**(a) shows that literature regarding industrial embodied intelligence remains limited from 1985 to 2018. However, the amount of literature began to grow significantly after 2018, and it peaked in 2024. **Fig. 1**(b) demonstrates that the amount of general research concerning "embodied intelligence" has grown steadily since 2002 and surged rapidly after 2020. The research regarding both topics grew slowly at first, and then rapidly increased. These trends are closely linked to recent breakthroughs in EIR, pre-trained models, and MLLMs. These techniques have enhanced the perception and cognition capabilities of robots, thereby causing them to be more adaptable to industrial settings. Geographically, the primary contributors to the industrial embodied intelligence research are China, the United States, Italy, the United Kingdom, and Germany. Notably, nearly 60% of the related publications originated in the computer science and engineering fields; this result reflects a growing trend of interdisciplinary integration. **Fig. 1**(e) and 1(f) list all the authors and affiliations that have published more than two articles in this field. These data suggest that a global research network surrounding industrial embodied intelligence has begun to form and is expanding rapidly.

[1] In this paper, EIIR may refer to either embodied intelligent industrial robotics (as a field of study) or an individual embodied intelligent industrial robot, depending on the context. When the meaning is unambiguous, the abbreviation is used without further distinction. EIR is used similarly.

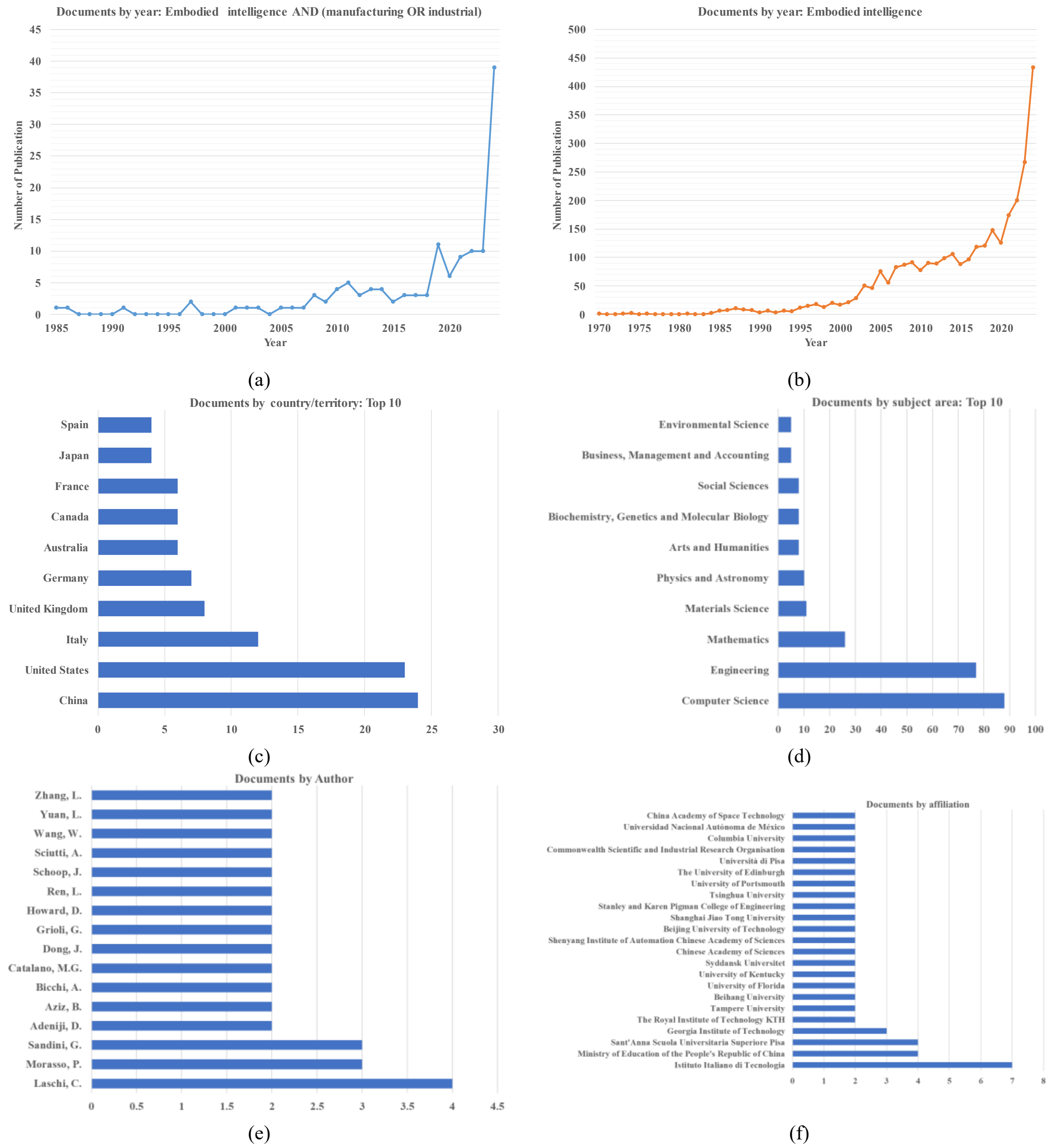

**Fig. 1.** Statistics obtained from Scopus (search keywords: '*embodied intelligence AND (manufacturing OR industrial)*': (a) number of papers by year ('*embodied intelligence AND (manufacturing OR industrial)*', 1985–2024), (b) number of papers by year ('*embodied intelligence*', 1970–2024), (c) number of papers by country or territory (top 10), (d) number of papers by subject area (top 10), (e) number of papers by author, and (f) number of papers by affiliation.

This review is different from existing work that has been performed in the fields of embodied intelligence and EIR. **Table 1** lists seven related review papers. The first four papers are foundational reviews that focus on embodied intelligence. Specifically, the first paper [13] emphasizes the interactions between morphology, action, perception, and learning in the agent architecture. The second paper [2] reviews recent techniques in four key areas: embodied perception, interaction, agent, and sim-to-real adaption. The third paper [6] evaluates the features and performance of simulators that have been used for embodied AI research, and particularly for tasks that involve visual exploration, navigation, and embodied question answering. In addition, the fourth paper [7] provides a macro-level perspective by examining embodied AI as one of several technologies for human-centered smart manufacturing. The last three papers explore the integration of large language models (LLMs) with robotics.

In contrast, this review makes three key contributions. First, it is the first review to focus on embodied intelligence in industrial robotics, summarizing the major challenges of EIR when applied to industrial scenarios and analyzing its deep reasons, and offering technical insights on EIIR. Second, a knowledge-driven EIIR technical framework is proposed to overcome the identified limitations, the massive literature is organized accordingly, and a concept-validation case study is given, too. Finally, the key challenges associated with the application of embodied intelligence to industrial settings are identified and future research directions are proposed.

**Table 1** Published embodied intelligence review papers.

| No. | Title | Venue | Year |
|---|---|---|---|
| 1 | Embodied intelligence: A synergy of morphology, action, perception and learning [13] | ACM Computing Surveys | 2025 |
| 2 | Aligning cyber space with physical world: A comprehensive survey on embodied AI [2] | IEEE/ASME Transactions on Mechatronics | 2025 |
| 3 | A survey of embodied AI: From simulators to research tasks [6] | IEEE Transactions on Emerging Topics in Computational Intelligence | 2022 |
| 4 | When embodied AI meets Industry 5.0: Human-centered smart manufacturing [7] | IEEE/CAA Journal of Automatica Sinica | 2025 |
| 5 | Large language models for robotics: Opportunities, challenges, and perspectives [14] | Journal of Automation and Intelligence | 2025 |
| 6 | A survey of robot intelligence with large language models [15] | Applied Sciences | 2024 |
| 7 | A survey on integration of large language models with intelligent robots [16] | Intelligent Service Robotics | 2024 |

The rest of the content is organized as follows. Section 2 introduces the definition and technical framework of EIIR. Section 3 elaborates upon the world model, which includes working-environment knowledge and operating-object knowledge. Section 4 focuses on techniques by which the high-level task planner can implement task decomposition, while Section 5 introduces techniques by which the low-level skill controller can achieve physical task execution. Section 6 provides an evaluation of the existing EIIR simulators and discusses the sim-to-real problem. Section 7 presents a case study of an EIIR-based flexible assembly system to illustrate the application of EIIR in industrial scenarios. Finally, challenges and future research directions are delivered in Section 8.

## 2 Background and framework of EIIR

In this section, a systematic overview of the development and interdisciplinary background of EIIR is first provided. Then, the existing EIR technical frameworks and their primary challenges are summarized. Finally, a new knowledge-driven EIIR technical framework tailored for industrial scenarios is proposed.

### 2.1 Industrial robotics: From automation to embodied intelligence

Since the advent of industrial robots in the 1960s, their definition, technical development, and classification was discussed and detailed by many seminal papers and reviews [17-21]. For example, Gao et al. [17] described the evolution from Robotics 1.0, where robots relied heavily on waypoint teaching to perform repetitive, labor-intensive tasks in structured environments, to Robotics 2.0, characterized by the integration of perception and feedback systems such as vision and force sensors. Singh et al. [18] emphasized that traditional industrial robots were suited to structured, pre-programmed tasks, whereas the next generation of sensor-equipped robots would be able to perform under less predictable and more dynamic conditions. Grau et al. [19] also described how robots in the 1990s largely operated in isolation without real environmental awareness, while the 2000s saw the emergence of sensor-based systems enabling limited human–robot collaboration. These perspectives collectively validate the transition from the automation era to the perception era, forming the historical foundation for our own extension. Here, our insights on IRs are mainly built upon the stage division proposed by Gao et al. in 2020 [17], but modified according to our specific focus and recent development trend in IR. As shown in **Fig. 2**, IRs can be roughly divided into three eras:

- *Automation era*: In the early stages of industrial robotics, the main advantage of IRs was their programmability. Research focused on robot bodies and their core components. These robots executed fixed tasks with predefined actions using hard-coded instructions. The pursued goal was higher efficiency and accuracy, while their flexibility was very limited.
- *Perception era*: With the advancement of sensors, machine vision, and deep learning, IRs gained strong perception and visual servo capabilities, which enabled a higher degree of skill-level flexibility. For example, due to their improved visual perception capabilities, the robots could manage parts that were not precisely positioned during loading and unloading tasks. However, they still lacked flexibility at the task level.
- *Embodied intelligence era*: The rapid development of EI techniques, such as MLLMs and world models, is bringing IR into a new era of EIR. In this new era, a single IR will become an industrial agent that is capable of perceiving, autonomous decision-making, and executing within its environment, much like a human worker. An IR will be able to complete a variety of tasks, such as loading, unloading, handling, palletizing, and assembly. Therefore, they will demonstrate true task-level flexibility, i.e. general intelligence in industrial scenarios.

Each industrial robotic era was built upon the technologies developed during the previous era. The authors believe that IRs are now entering the early stages of the e*mbodied intelligence era*. A key future trend is the integration of embodied intelligence with industrial robotics to form a new research field EIIR, which would be applied universally across diverse industrial systems in the near future.

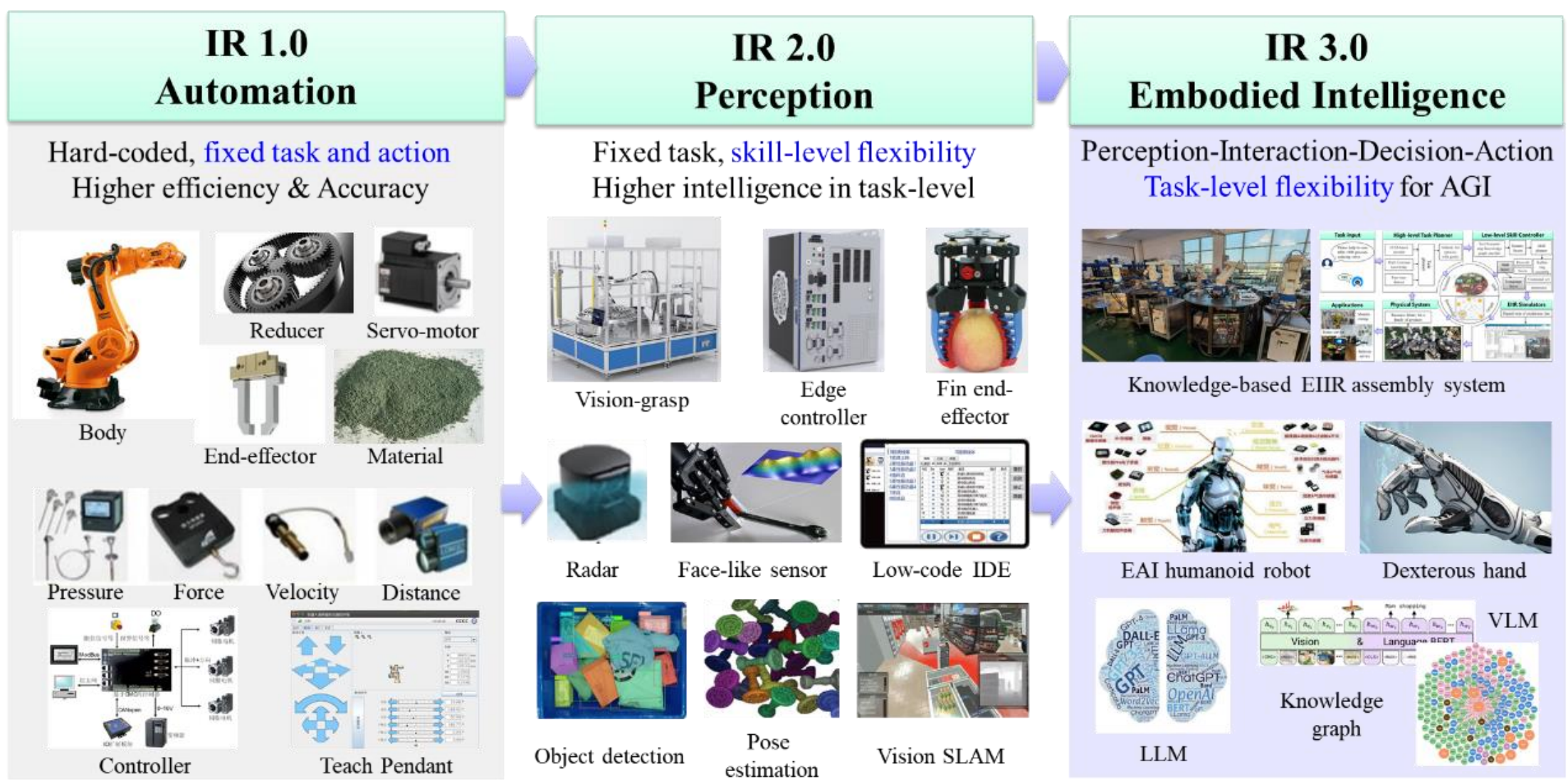

**Fig. 2.** The three eras of industrial robotics.

Here we make the first attempt to systematically organize the concepts related to embodied intelligence and industrial robotics. The relationships between artificial intelligence (AI), embodied intelligence (EI), embodied intelligent robotics (EIR), and embodied intelligent industrial robotics (EIIR) are summarized in **Fig. 3**. AI primarily consists of three technical schools: symbolism, connectionism, and behaviorism [22]. EI represents the frontier of behaviorism. It primarily studies agents that can perceive, make decisions, and interact with the environment. It emphasizes that an agent can become smarter through environmental interactions and not just by symbolic computation; thus, it aligns with the embodied Turing test proposed by Turing in 1950 [23]. EI contains two primary categories: virtual agents and physical agents [2, 13, 24]. Of the physical agents, robots are the most suitable carriers, leading to the emergence of EIR. An EIR uses multi-modal sensors to perceive its environment, uses cognitive models to achieve dynamic decision-making, and controls physical actuators that interact with objects and enable the completion of complex tasks. The specific forms of an EIR are heavily dependent upon its application domain. Existing EIR forms mainly include humanoid robots, quadruped robots, mobile robots, industrial robots, and service robots [2]. Within this framework, industrial robots represent a specific EIR form that is designed for industrial applications, referred as EIIR. Therefore, EIIR focuses on industrial agents that are equipped with independent perception, decision-making, and execution capabilities that are tailored for industrial environments, data, and tasks. As with EI, industrial embodied intelligence can also be divided into two categories: virtual and physical. Virtual industrial agents can exist in robot or production-line simulators (which are discussed in Section 6). EIIR focuses on typical physical industrial agents to achieve task-level flexibility.

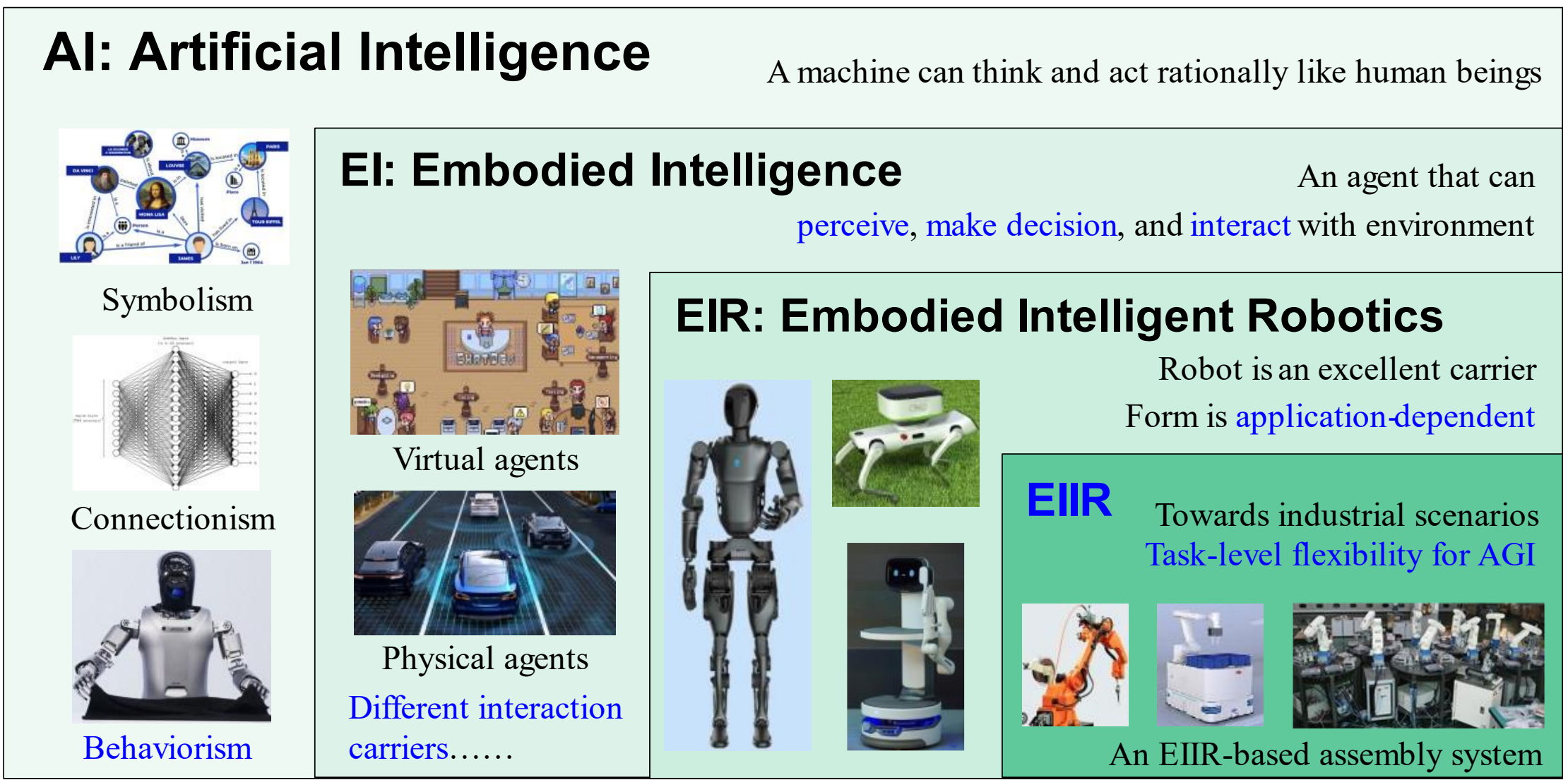

**Fig. 3.** Relationships between AI, EI, EIR, and EIIR.

## 2.2 Overview of the EIIR framework

The existing EIR frameworks can be broadly classified into two categories: hierarchical architecture and end-to-end architecture. The hierarchical architecture usually has a dual-system design inspired by human brain and cerebellum, as seen in example like humanoid robot of Figure AI [25]. It typically consists of two primary sub-systems: a high-level planner and a low-level controller [2]. The high-level planner uses MLLMs to process text (e.g., "take an apple to my room") and visual inputs (such as image captions or scene graphs). It then uses semantic reasoning to decompose the abstract task into a sequence of executable subtasks (e.g., go to the kitchen → find the fridge → open the fridge → ...). The low-level controller manages the task execution. It uses perception models (which can be used to estimate the position and orientation of the apple, for example) and manipulation models (which can be used to generate manipulator actions, for example) to perform each subtask, and it is guided by real-time sensor feedback. Simulators can also be integrated into this framework to train and test agents in diverse high-fidelity virtual environments. This virtual–real integration approach is less expensive than physical trial-and-error methods and it supports system self-improvement through continuous feedback from environmental interactions.

The end-to-end-architecture integrates vision, language, and action into a single model, known as a vision–language–action (VLA) model. Such a model is used to directly model the full closed loop that includes perception, decision, and action. Representative studies belong to this category includes Google's RT series [3, 26] and OpenVLA [27]. This architecture typically consists of three core modules: a multi-modal input processing module, a cross-modal fusion module, and an action decoding module. First, the VLA model obtains three types of inputs: visual data (images or video), language text, and action data (such as past motion trajectories). Next, using a cross-attention mechanism, the model aligns visual features, language embeddings, and action representations within a shared semantic space. This alignment enables the model to understand the relationships between different modalities and effectively combine their information. Finally, the combined multi-modal representation is passed to an action decoder, which generates either continuous control signals (e.g., joint angles for a robotic arm) or discrete action sequences (e.g., navigation paths or manipulation steps).

However, there are significant differences in the technical requirements raised from industrial and daily-life scenarios. In industrial scenarios, a technology is usually required to have much higher: efficiency, accuracy, reliability, and safety. These are important gaps for existing EIR techniques when transferred from daily-life scenarios to industrial scenarios. Some academic researchers [5, 7] and commercial companies (e.g. Figure AI, ZhiYuan Robot) have realized these challenges. First, current MLLMs are constrained by limited industrial knowledge, insufficient incorporation of process information, and a general lack of industrial constraints [9-11, 28]. As a result, current large models pretrained on internet-based textual corpora struggle to generate task planning schemes that satisfy the strict constraints inherent in industrial operations. More specifically, along the industrial task-planning chain, these planning limitations can be summarized as failures in: (1) *process-information structuring*, where MLLMs fail to reliably transform process cards, assembly manuals, technical specifications into clear operation nodes, dependency relations, and key process parameters; (2) *operation-dependency reasoning*, where the generated sequence may appear reasonable in natural language but does not respect the actual process logic among workpieces, intermediate states, and process flows; and (3) *constraint-compliant plan generation*, i.e., overlooking geometric, mechanical, safety, or quality constraints required for executable industrial operations. Second, existing EIRs, which are primarily trained for daily-life scenarios, lack support for industrial skills and their corresponding control languages. As a result, they are unable to execute essential operations such as press-fitting or screwing, which are fundamental in industrial tasks. Finally, existing embodied simulators are not designed for industrial embodied intelligence. They struggle to replicate the complexity of industrial environments, such as multi-type sensors and heterogeneous controllers like programmable logic controllers (PLCs). This limits industrial agents' ability to train and perform virtual commissioning in the virtual industrial environment.

To address these challenges, a knowledge-driven EIIR technical framework was developed during this study. as illustrated in **Fig. 4**. It is tailored to the needs of industrial scenarios, data, and tasks. It consists of five components: a world model, a high-level task planner, a low-level skill controller, a simulator, and a physical system. The world model used here refers to a symbolic formulation rather than a generative world model that synthesizes visual scenes [29]. The world model, which serves as the primary knowledge source for the agent, is at the center of this framework. It provides general knowledge, working-environment knowledge, and operating-object knowledge. The general knowledge provides the semantic foundation of the LLM for the interpretation of natural language tasks. The working-environment knowledge takes the form of a semantic map of the production line in which the equipment orientations, operable boundaries, and other environmental constraints are dynamically marked. The operating-object knowledge takes the form of a domain-specific knowledge graph that structurally stores product processes and parameters, thereby enabling the planner to generate subtask sequences that align with the industrial specifications. When a new product is introduced, its process, part topology, and parameter specifications can be automatically extracted from process documents and computer-aided design (CAD) models into the operating-object knowledge graph using LLM with industrial task-oriented fine-tuning.

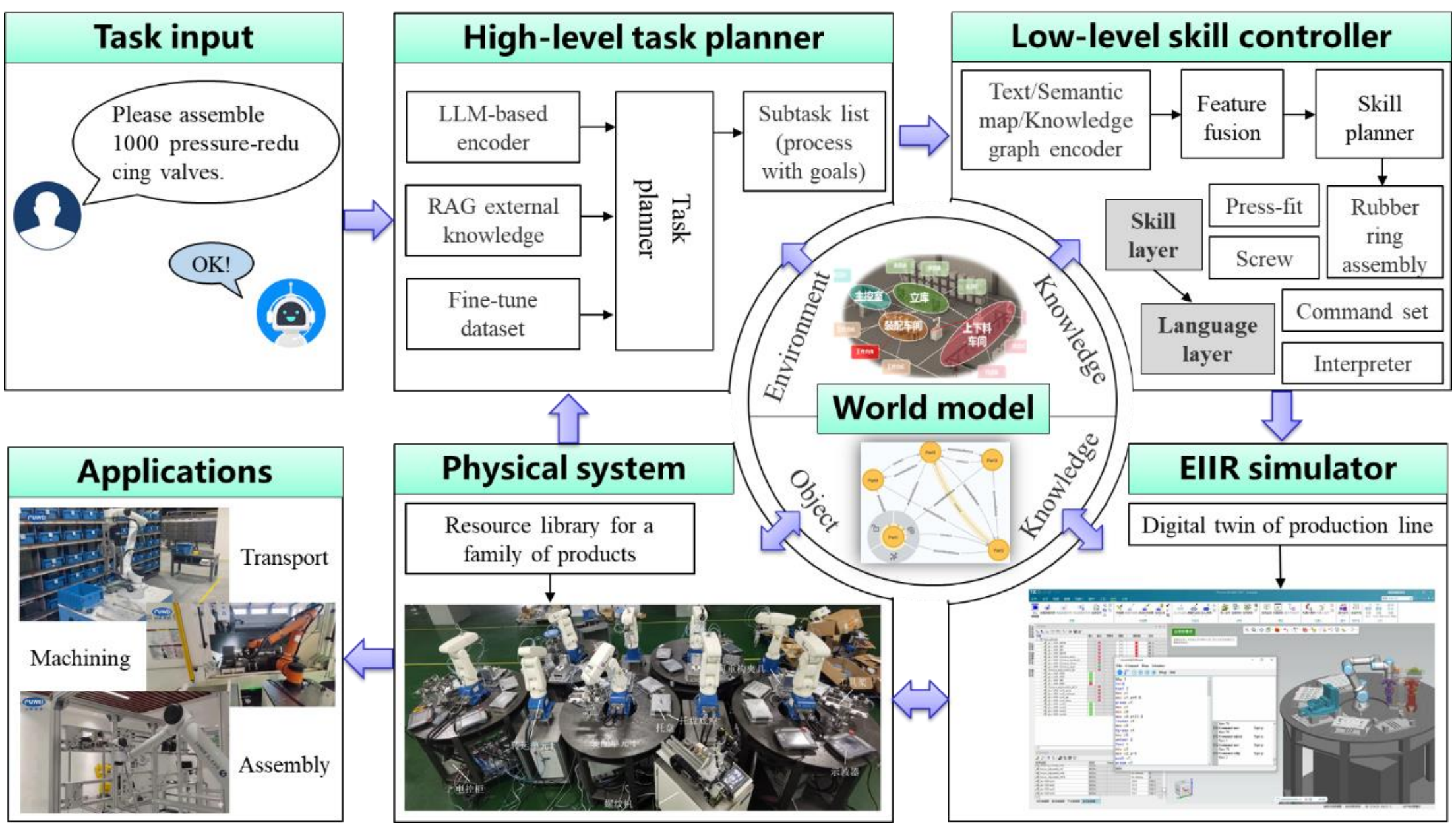

**Fig. 4.** A knowledge-driven EIIR technical framework.

With this EIIR framework, when the user gives a command, such as "Please assemble 1,000 pressure-reducing valves.", the high-level task planner parses the semantics through an LLM-based encoder and integrates external knowledge to disassemble the task into a series of subtasks. Next, the low-level skill controller maps the abstract subtasks to physical operations, and the skill layer can call the pre-defined industrial skills in the library and convert them into device-independent standardized instructions in the language layer. An interpreter is then used to dynamically convert these instructions into protocol instructions for various target controllers, which drive the device to execute the tasks. Since the low-level execution depends on deterministic industrial skills, the same skill set can be reused under different product configurations. The EIIR simulator then builds a production line-level digital twin to generate a dataset of manufacturing conditions in a virtual environment. It also achieves cross-platform collaborative simulation that includes the robot and the PLC by using middleware, such as robot operating system (ROS) or OPC server, to verify the correctness of the generated instructions. Finally, the virtual–real production line operation is completed in the physical system.

It is noteworthy that rather than executing in a one-shot manner, the operation proceeds as a perception-decision-action loop that aligns with EI principles. The simulator and the physical system form a bidirectional digital-twin loop in which robot data, PLC states, and vision signals flow back to the simulator during execution, while virtual sensors and geometric/kinematic models allow the simulator to infer additional production states beyond those directly sensed by hardware. These perceptual signals continuously update the world model, including both operating-object states and working-environment states, enabling the task planner and skill controller to re-evaluate or adjust strategies when deviations occur.

The next five sections of this paper focus on the world model, the high-level task planner, the low-level skill controller, the EIIR simulator and the physical system as a case study of this knowledge-driven EIIR framework. These five sections summarize the existing related techniques and explain how the components work together. They therefore provide a reference for scholars or engineers who are tasked with the selection of appropriate EIIR techniques.

# 3 World models for industrial scenarios

To operate effectively in industrial scenarios, EIIRs require a world model—an internal representation that provides agents with general knowledge (LLM semantic understanding capability), working-environment knowledge, and operating-object knowledge. These knowledges enable natural human-machine communication and supply intensive knowledge for advanced techniques development in other modules, to make robots working more efficiently, accurately, reliably, and safely in industrial scenarios possible. The following sections detail the methods of representing the working-environment and operating-object knowledge.

## 3.1 Working-environment knowledge

Working-environment knowledge refers to the spatial, topological, and semantic information of the surrounding environment that is necessary for robots to perform autonomous navigation, obstacle avoidance, and object manipulation in industrial scenarios. This includes the location of equipment, operable boundaries, and other physical constraints that define how the robot can interact with its environment. Over time, the representation of working-environment knowledge has evolved through several stages:

1) *Geometric map*: A geometric map is a low-level spatial representation that encodes the geometric structure of an environment, typically in terms of obstacles, free space, and physical boundaries. Such maps are most commonly generated using simultaneous localization and mapping (SLAM) techniques, where the robot simultaneously estimates its own position and incrementally builds a map of the environment through sensor observations [30]. However, a key limitation of geometric maps is their lack of semantic information. They only capture geometric and spatial features but do not encode the identity or meaning of objects. For example, while the map can indicate the location of an obstacle, it cannot distinguish whether the obstacle is a table, a robot, or a machine. This restricts the robot's ability to perform high-level reasoning and semantic task execution.

2) *Scene graph*: A scene graph is a semantic representation that depicts objects within a scene along with their categories and relationships [31]. In a scene graph, each node represents an object, while each edge captures a relationship between two objects, such as spatial relations (e.g., "on top of", "next to"). Scene graphs serve as a semantic context guide in tasks such as visual grounding and task planning. They enable robots to dynamically infer and update the semantic structure of their surroundings, especially in cluttered or changing environments. It is worth noting that scene graphs, by focusing solely on objects and their spatial relations without incorporating the rich attributes, diverse relation types, and reasoning rules of knowledge graphs, offer a lightweight graph structure that enables faster real-time updates and reasoning. However, a key limitation of scene graphs is the absence of detailed geometric or spatial layout information. While object locations (e.g., 2D/3D coordinates) can sometimes be included within nodes, scene graphs do not reconstruct the physical geometry of the environment.

3) *Semantic map*: A semantic map is a comprehensive representation that integrates both geometric and semantic information about the environment [32]. Semantic maps enable robots to achieve both spatial reasoning and semantic reasoning within a unified framework. This dual capability allows robots to perform high-level autonomous behaviors, such as goal-directed navigation, semantic-aware manipulation, and context-driven task execution.

To construct a semantic map, researchers first use various scene reconstruction methods to obtain the geometric information of the scene [33, 34]. Then, deep-learning based object detection methods [35] and knowledge graph techniques [36] are used to extract object-related semantic information, such as categories, relationship semantics, and functional semantics. With the development of LLMs and vision–language models (VLMs), a significant amount of new work regarding semantic maps of high-dimensional scenes has been performed. In this section, LLMs are used as a distinguishing point so that the construction methods can be classified as either LLM-based methods or non-LLM methods. The characteristics of both types of methods are presented in **Table 2**.

**Table 2** Comparison between the two types of semantic map construction methods.

| Characteristic | Non-LLM methods | LLM-based methods |
| --- | --- | --- |
| **Semantic openness** | Closed sets | Open vocabulary |
| **Dynamic adaptation** | Dependent upon geometric updates | Real-time reasoning |
| **Calculation load** | Low, friendly deployment | High, LLM collaboration |
| **Generalization** | Fine-tuned according to the scene | Zero-shot cross-scene migration |

The non-LLM methods generally use a graph neural network (GNN) to combine the prior features in various 3D semantic scenes for feature extraction. Finally, the features are used to predict the semantic label of each object and the semantic relationships between the objects. Wald et al. [37] proposed a learning method of obtaining a semantic map from the point cloud of a scene. This method is based on the PointNet and uses a graph convolution network (GCN) to generate the semantic map. These scholars also introduced a semi-automatic dataset, which was based on this task, that contained semantic maps with sufficient semantic information. This work and dataset serve as a good foundation for the development of semantic map generation. To address the incrementalization and real-time requirements of robot-application scenarios, Wu et al. [38] proposed a method, referred to as SceneGraphFusion, in which RGB-D video sequences were used to incrementally construct a semantic map. The method uses a GNN to aggregate PointNet features from the components of the original scene. In addition, the authors proposed a new attention mechanism, which enables effective frame-by-frame reconstruction of the scene in the presence of incomplete data and missing graph data. Subsequently, to address the high computing power required by the dense point cloud during semantic map construction, they optimized the algorithm discussed above and proposed MonoSSG [39]. Based on multi-modal features, such as sparse point clouds and scene images, the algorithm used multi-view and set features to aggregate GNNs and predict the semantic map. This method significantly improved the semantic map construction speed while maintaining good accuracy. For indoor scenes with complex structures and dynamic scenes with pedestrians, Rosinol et al. [40] proposed a method of constructing a directed 3D dynamic semantic map. The nodes in the map represent entities in the scene (such as objects, walls, and rooms), while the edges represent the relationships between the nodes (such as inclusion and proximity). The map also includes mobile agents (such as humans and robots) and operable information (such as spatial–temporal relationships and topological relationships at various abstraction levels) to support planning and decision-making. To address the problem of semantic map construction during real-time robot-perception processes, Hughes et al. [41] proposed Hydra, which was a real-time semantic map construction algorithm. A Euclidean signed distance field (ESDF) is used to reconstruct the scene perceived by the robot. In addition, the semantic map constructed by the ESDF is divided into hierarchical rooms so that a multi-level semantic map can be built. In this method, a loopback detection and global optimization algorithm is also constructed for the map; this algorithm can achieve real-time and efficient semantic map construction for the robot. In general, non-LLM semantic map construction algorithms usually use point clouds and images as the feature inputs. The model architecture uses a GNN as the intermediate framework for feature aggregation, and the semantic map construction is organized according to the spatial topology. This type of algorithm has a fast-reasoning speed, is suitable for end-to-end deployment, and can satisfy the dynamic scene-construction requirements; however, its semantic-feature dimension is limited. Therefore, it is difficult to obtain the complex semantic information in the scene when using this type of algorithm.

The development of LLMs has brought a new research perspective to the extraction and generation of semantics. An MLLM that is integrated with vision can perceive and summarize many kinds of semantic information in a scene image. In addition, the large model can further conduct retrieval, reasoning, and planning tasks according to the summarized semantic information; thus, the semantic perception capability for the actual scenes can be significantly enhanced. Chang et al. [42] proposed an open vocabulary-oriented semantic map construction framework that obtained the connections between various entities in the form of natural-language text output. Different from traditional semantic-based object localization methods, this framework supports context-aware entity localization; thus, it allows entity location-based queries, such as "pick up a cup on the kitchen table" or "navigate to the sofa where someone is sitting." Unlike existing semantic map research, this approach supports free text inputs and open vocabulary queries.

To address the problem of single-modal labels of semantic maps, Jatavallabhula et al. [43] proposed a multi-modal semantic map construction method, which was referred to as Conceptfusion. This method can solve the closed-set restriction of the existing semantic map concept reasoning and expand the semantic retrieval to the open set of natural semantics. In addition, the semantic maps constructed by this method contain multi-modal semantic attributes. The method can retrieve objects from the map using language, image, audio, and 3D-geometry inputs. Conceptfusion uses the open-set capability of the foundation model that is pre-trained on internet scale data to infer concepts of different modes. This method has zero-shot characteristics, does not require any additional training or fine-tuning, and can better retain the long-tail concept; thus, it is superior to the supervised method.

To address the complex and diverse difficulties associated with using RGB-D video sequences to construct semantic maps of large scenes, Gu et al. [44] proposed a method that utilized an LLM. This method employs a 2D detection and segmentation model, and it integrates the output of detection results into 3D through multi-view RGBD sequences. This method also has zero-shot characteristics, and it can construct a semantic map without collecting a large number of 3D datasets or fine-tuning models. Experiments demonstrated that this method could support the prompt assignment of user inputs, as well as complex reasoning that integrated the understanding of spatial and semantic concepts (which is a downstream planning task).

To address the extremely complex spatial-description problem of multi-story building navigation, Werby et al. [45] proposed a semantic map construction method, referred to as HOV-SG, for multi-story and multi-room navigation tasks. First, an open-vocabulary visual foundation model was used to construct a 3D open-vocabulary semantic map. Then, the floors and rooms were divided in the map, and the room names and types were determined. Finally, the results were used to construct a 3D multi-level map. The main features of the method are its ability to represent multi-story buildings and provide semantic connections for robots in buildings. This method produces very good experimental results when used for a long-distance, multi-story building navigation task.

The complexity and diversity of the semantic information of outdoor scenes is a primary cause of the application restrictions that are associated with semantic map construction techniques. To solve this problem, Strader et al. [46] proposed an ontology-based indoor and outdoor general semantic map construction technique. First, the author proposed a method of establishing the spatial ontology and defined the concepts and relationships related to the operation of indoor and outdoor robots. In particular, the author used an LLM to build a basic semantic ontology, which significantly reduced the manual-annotation workload. Next, the author used a logic tensor network (LTN) to construct a semantic map based on the spatial ontology. The logic rules and axioms that were added to the LLM provided additional monitoring signals during training, and thus reduced the need for labeled data. This method provided more accurate predictions, and it even predicted concepts that were not seen during training.

During semantic map construction, LLMs act as semantic-related operators by performing various tasks, such as semantic extraction, reasoning, and classification; however, the existing work has placed more focus on semantic extraction and 3D reconstruction based on visual perceptions. For robots, understanding abstract spatial semantics is an important prerequisite to perceiving the physical world. LLMs build intuitive tools that robots can use to perceive the physical world; thus, they enable the in-depth semantic understanding and operation of robots. Semantic maps are the intermediate expressions and elements in this type of operation.

While semantic map has been extensively studied in general embodied AI, its deployment in industrial environments is still at an early stage. There exist isolated applications that demonstrate the potential of semantic map in industrial settings, such as the construction of semantic maps for human-robot collaborative manufacturing, where geometric and semantic information is integrated to support robot interaction on factory floors [47]. Similarly, a LiDAR-RGBD-odometry fusion system has been applied to a desalination plant to automatically detect and label infrastructure components, illustrating how scene understanding can benefit industrial inspection and navigation tasks [48]. However, these examples represent the application of general techniques to industrial scenarios, rather than the development of domain-specific frameworks. A fundamental gap lies in the lack of industrial foundation models capable of robustly supporting industrial object recognition, relational reasoning, and functional semantic understanding. Therefore, the combination of semantic map and industrial embodied intelligence remains largely underexplored.

## 3.2 Operating-object knowledge

Operating-object knowledge refers to the domain-specific knowledge about industrial products, processes, and associated resources, which plays a crucial role in enabling robots to perform task planning and execution in accordance with industrial constraints. This knowledge becomes even more critical as manufacturing shifts toward greater flexibility and customization because it requires EIIR to manage mixed-line production that involves multiple product types and small batch sizes. In the context of robotics, operating-object knowledge is predominantly represented using two forms: ontology and knowledge graph [49]. Ontology is a formal abstraction of domain knowledge, primarily designed to express, share, and reuse knowledge. It provides a well-structured, hierarchical description of concepts within a specific domain. With the evolution of ontology, knowledge graph has emerged as a more advanced representation, integrating a pattern layer (conceptual schema defined by ontologies) and a data layer (instance-level information) [50]. Furthermore, knowledge graph offers stronger reasoning capabilities by combining traditional ontology-based symbolic reasoning with flexible graph database querying, enabling more efficient knowledge retrieval and inference in dynamic industrial environments.

Researchers have developed industrial knowledge graphs of varying complexities for industrial scenarios; these graphs are centered around three fundamental elements: products, processes, and resources [51]. **Table 3** summarizes the key characteristics of this body of work. For instance, Bharadwaj [52] hierarchically structured the product information. The author divided items into assemblies, sub-assemblies, and parts, and then stored this structure in a knowledge graph. Chen [53] proposed a knowledge graph-based assembly information model, referred to as KGAM, that integrated product data extracted from CAD models with process data obtained from technical documents. This unified representation allows engineers to query the manufacturing process and its attributes for specific products, thereby enabling a semantic link between the products and the process. Shi [54, 55] extended this approach by building an industrial knowledge graph that incorporated resources alongside the products and processes (see **Fig. 5**). This resource-centric graph aids in the management of design assets and facilitates asset reuse in future projects. In summary, industrial knowledge graphs offer clear and structured representations of the entities and relationships in industrial scenarios. They effectively address data heterogeneity across the design, planning, and production stages, and they serve as a critical knowledge bases for autonomous decision-making in industrial embodied intelligence.

**Table 3** Comparison of the industrial knowledge graphs found in published literature.

| **Literature** | **Elements** | | | **Applications** | | |
|---|---|---|---|---|---|---|
| | | | | High-level | | Low-level |
| | Product | Process | Resource | Sequence planning | Resource allocation | Action matching |
| Bharadwaj [52] | √ | | | | | |
| Jia [56] | √ | | | | | |
| Liu [57] | √ | | | √ | | |
| Chen [53] | √ | √ | | | | |
| Hu [58] | √ | √ | | | | |
| Zhou [59] | √ | √ | | √ | | |
| Xiao [60] | √ | √ | | √ | | |
| Shi [54, 55] | √ | √ | √ | | √ | |
| Järvenpää [61, 62] | √ | √ | √ | | √ | |
| Mo [63, 64] | √ | √ | √ | | √ | √ |
| Zhong [65] | | √ | √ | | | √ |
| Xu [66] | √ | √ | √ | √ | √ | √ |

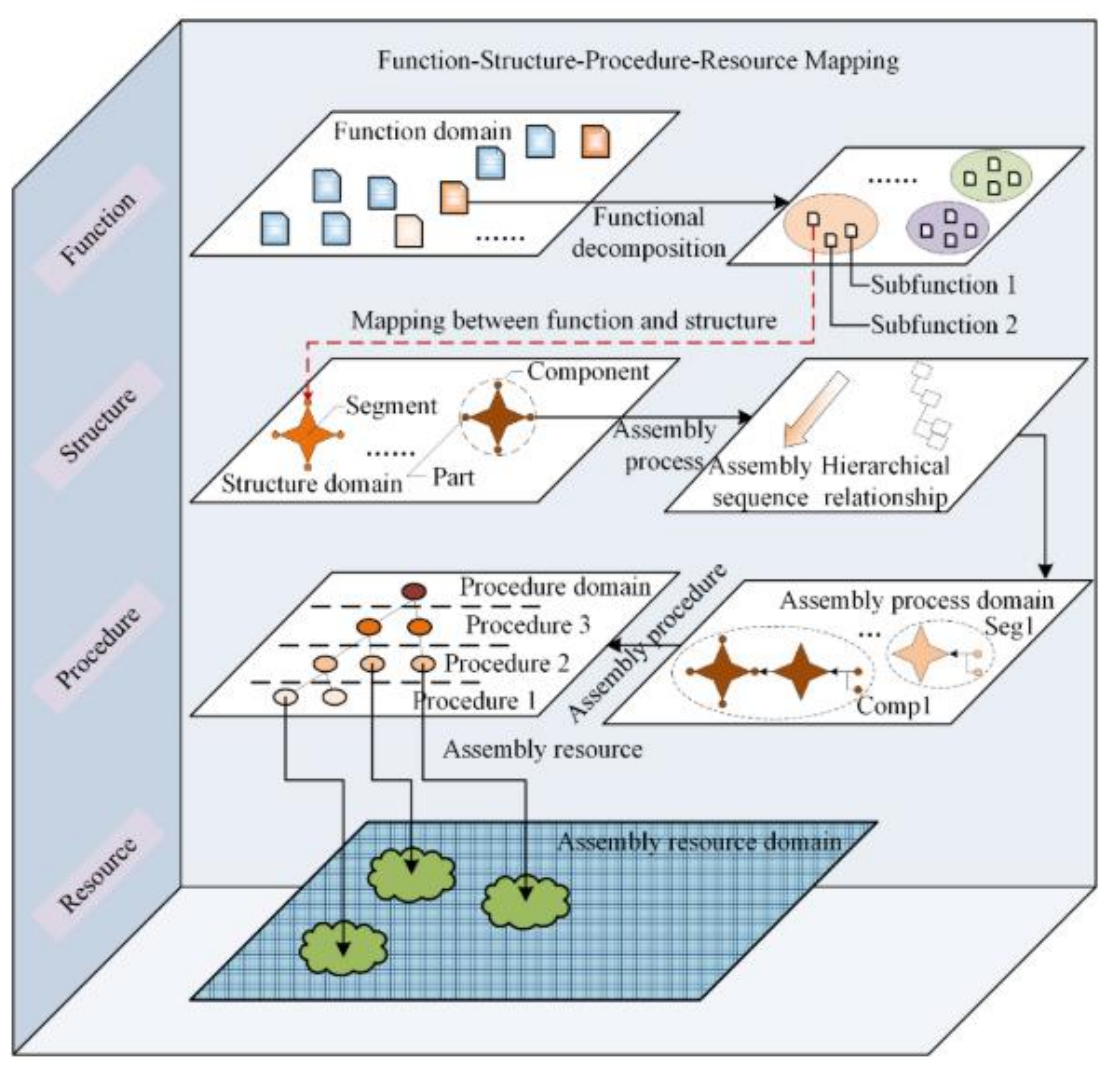

**Fig. 5.** Industrial knowledge graph that combines products, processes, and resources [54].

Using an industrial knowledge graph, an EI-powered industrial system can support both high-level and low-level decision-making tasks that include sequence planning, resource selection and allocation, and action matching. First, sequence planning is conducted using the knowledge, rules, and algorithms embedded in the product and process knowledge graph. For example, Liu [57] developed a three-layer assembly information model that represented product data and supported planning according to instantiated knowledge. Second, to improve resource reusability, Järvenpää [61, 62] defined a conceptual resource structure. This structure integrates the product, process, and resource information, thereby enabling process–resource matching. Similarly, in response to changing requirements, Mo [63, 64] built a knowledge graph that uses real data from OmniFactory to support both resource selection and system reconfiguration. Finally, knowledge graphs also support action matching at the execution level. Zhong [65] stored robot skills in a knowledge graph, thereby allowing the system to match suitable skills with the input task. These skills are then decomposed into basic actions for execution. Notably, Xu [66] present a comprehensive knowledge-graph-based assembly task planning framework that integrates sequence planning, resource allocation, and execution-level skill mapping within a unified industrial system.

# 4 High-level task planner

The high-level task planner is located in the top level of the EIIR framework. It receives tasks described by users in natural language and converts them into subtasks to guide the specific actions of robots [2]. This process enables the robots to understand natural language, which enhances the convenience of robot control by non-professional operators. In this section, general task-planning techniques based on general knowledge are first introduced. Next, task-planning techniques for industrial scenarios are presented along with their specifications and constraints.

## 4.1 General task planning

General task planning uses general knowledge to manage tasks in unstructured settings, such as shopping malls and restaurants. Its goal is to divide tasks into sequential subtasks by means of natural language inputs. By combining visual perception and multi-modal information processing, robots can adapt to environmental changes and complete tasks in real time. For instance, if given the command "Please help me get an apple," a robot may split this task into these subtasks: "Find the apple," "Go to the location of the apple," "Grab the apple," "Navigate to the dining table," and "Place the apple on the table without touching any other objects." Meanwhile, the robot can use vision to collect relevant information from the scene. If the scene changes, the robot can automatically update and adapt according to the visual signals or sensor data. This process addresses the generalization problem associated with traditional planning methods. In this study, according to the planning methods and input modes, the existing general task-planning methods were categorized as either **LLM-based methods** or **VLA-based methods**.

### 4.1.1 LLM-based methods

LLM-based general task-planning methods use tasks that are described using natural language as inputs. They utilize the reasoning capabilities of LLMs to decompose complex tasks. The core features of this technique are the powerful language understanding, reasoning, and generation capabilities of LLMs. The use of specific modular designs can further improve the planning capabilities of LLMs in complex environments. Researchers have introduced various auxiliary modules, such as the prompt, visual module, and security module that are described in **Table 4**. These modules were added to improve the LLM adaptability and enable more efficient task planning.

1) **LLM + prompt.** Inputting specific prompts to an LLM can significantly enhance its ability to sub-divide the input tasks. PROGPROMPT [67] utilizes a programmed prompt structure. It guides LLM to generate task plans, combined with operable objects in the environment and sample programs provided to LLM. The LLM-GROP [68] and G-PlanET [69] symbolize environmental information and store it in data or symbol form, thereby allowing the LLM to use the environmental information for planning. The LLM-State [70] treats the LLM as an attention mechanism, a state estimator, and a strategy generator. It addresses long-sight distance issues in open worlds and allows the LLM to adjust the task-planning process in real time according to the scene information. The GRID [71] is a graph-based robot task resolver. It uses scene graphs rather than images to perceive global scene information and iteratively plans the subtasks for a given task.

2) **LLM + visual module.** The addition of a visual module to the LLM enables the model to perceive environmental information that it can use for task planning. When only an LLM is used, the agent cannot sense the surroundings in real time. For task planning in complex environments, the LLM-Planner [72] uses a visual module to collect physical environmental information. It performs dynamic replanning if a task fails or times out. The TaPA [73] did not directly use existing large models. Instead, it built a planning dataset to fine-tune the LLaMA-7B, which enhanced the task-planning success rate. The ViLaIn [74] integrates the Grounding-DINO scene-detection module. It converts scene information into the planning domain definition language (PDDL) format. The initial PDDL state was generated by combining the BLIP-2 and GPT-4 models. The ViLaIn also introduces corrective re-prompting error feedback and chain-of-thought (CoT) mechanisms. These mechanisms improve the granularity and accuracy of the generated tasks. Liu et al. [75] added high-quality teaching cases to visual information, thereby enhancing the reasoning capability of the robot for complex problems. Prompt2Act [76] further maps multi-modal prompts, including language instructions and goal images, into sequences of actionable robotic steps using large foundation models, providing another prompt-driven route for grounded robotic planning.

3) **LLM + security module.** The addition of a security module to the LLM ensures the safety and reliability of the generated plan. The LLM may be unaware of certain details in the actual scene, and this ignorance could lead to dangerous robot actions. The CLSS [77] includes a cross-layer sequence supervision mechanism. Using linear temporal logic (LTL) syntax, it expresses safety constraints and violations that are detected during the task and motion planning processes, then it corrects them. The SafeAgentBench dataset [78] was developed to assess the safety of existing planning methods. It evaluates methods and determines whether they are safe and reliable. The dataset includes 750 tasks, 10 hazards, and three task types. The authors tested eight LLM-based agents and evaluated them using the rejection, success, and execution rates. Their results showed that the security and stability of current agents were still weak. The Safe Planner framework [79] incorporates a safety module, which endows the LLM with safety awareness. It uses a multi-head neural network to predict the execution-skill safety. The ROBOGUARD [80] combines high-level safety rules with the environmental context of the robot. It uses the CoT reasoning mechanism to create strict and adaptable safety rules. Its contextual grounding module uses a root-of-trust LLM to transform abstract safety rules into absolute LTL formulas for inference.

**Table 4** Classification of the existing LLM-based planning methods.

| Category | Method | Base model | Evaluation | Core principle |
|---|---|---|---|---|
| LLM + prompt | PROGPROMPT [67] | GPT-3 | SR, Exec, GCR | Improves LLM understanding of human commands and robot tasks. |
| | LLM-GROP [68] | GPT-3 | UR | |
| | G-PlanET [69] | BART | CIDEr, SPICE, KAS | |
| | LLM-State [70] | GPT-4 | ACC | |
| | GRID [71] | INSTRUCTOR | SR | |
| LLM + visual module | LLM-Planner [72] | BERT | ACC | Uses visual information to improve the environmental |
| | TaPA [73] | LLaMA-7B | SR | |
| | ViLaIn [74] | GPT-4 | Rsyntax, Rplan, Rpart, Rall | perception capabilities of robots during task execution. |
| | H. Liu et al. [75] | GPT-4 | SR, Exec | |
| | CLEAR [81] | GPT-4, GPT-3.5, LLaMA2 | SR | |
| LLM + security module | CLSS [77] | GPT-4 | SFR, SR, Exec | Improves the plan-generation and task-execution safety. |
| | SafeAgentBench [78] | GPT-4 | Rej, SR, ER | |
| | Z.Yang et al. [82] | GPT-4 | SFR, SR | |
| | ROBOGUARD [80] | GPT-4o | ASR | |
| | Safe Planner [79] | GPT-4 | Collisions, SR | |

**Table 4** also lists the evaluation metrics that were used in the studies. The success rate (SR) was defined as the ratio of the number of tasks that the robot successfully completed to the total number of tasks that were executed. The executability (Exec) was defined as the ratio of tasks that the robot could perform to the total number of tasks that were generated. The accuracy (ACC) refers to the accuracy with which the robot performed the tasks, and it is used to measure the task-completion quality. The rejection rate (Rej) is defined as the ratio of the number of times the robot refused a dangerous task to the total number of tasks that were generated. The execution rate (ER) is defined as the ratio of tasks that were actually executed by the robot to the total number of tasks that were generated. The safety rate (SFR) is defined as the ratio of the number of times the robot performed a task without dangerous behavior to the total number of tasks that were executed. The user rating (UR) refers to a subjective evaluation by the user of the robot task performance; it is expressed as scores or grades. The key action score (KAS) is used to evaluate the robot performance with respect to key actions. Consensus-based image description evaluation (CIDEr) and semantic propositional image caption evaluation (SPICE) are indicators that are used to evaluate image captioning tasks. They evaluate the quality of generated text by comparing it with reference descriptions. They are used in robotics applications to assess the accuracy of natural language descriptions or instructions that are generated by robots. The collisions indicator reports the number of collisions that occurred during task execution. The Rsyntax indicator is used to evaluate the syntax correctness of the generated planning descriptions (PDs) by reporting the proportion of grammatically correct PDs. The Rplan indicator is defined as the proportion of PDs with effective plans. The Rpart and Rall indicators are used to evaluate the similarity between a generated PD and the ground truth.

In recent years, more research has focused on high-level robotic task planning. However, robotic task planning is merely the first step in robotic task execution. An embodied agent needs both a high-level task-planning capability and a corresponding low-level action controller. The emergence of VLA techniques has combined high-level planners with low-level controllers to directly generate specific robot actions. The differences and connections between VLA techniques and LLM-based task-planning methods are further described next.

#### 4.1.2 VLA-based methods

The general VLA-based method considers both visual information and natural language input during the task-planning process. Typical VLA architectures are shown in **Fig. 6** [83]. VLA models can directly convert natural language inputs into specific actions that robots can execute. Generally, the LLM is only a part of the VLA model. There are three primary differences between VLA-based and LLM-based task-planning methods:

- **Input-mode difference:** LLMs only accept language as inputs. Thus, when a robot performs task planning using an LLM, it must combine the LLM with other modules to perceive environmental information. In contrast, VLA models integrate vision, language, and action. They can directly utilize visual information to enhance the ability of the robot to understand the environment.
- **Architecture difference:** An LLM primarily consists of a language encoder and a decoder. In contrast, VLA architecture is more complex. It includes a visual encoder, a language encoder, and an action decoder. This architecture allows the VLA model to directly integrate visual and textual information and to generate specific actions.
- **Specific task-planning difference:** LLM-based task-planning methods can only generate subtask sequences. They must work with low-level controllers to interact with the physical world. VLA models, however, directly translate text and visual information into specific actions, thereby enabling better environmental interaction.

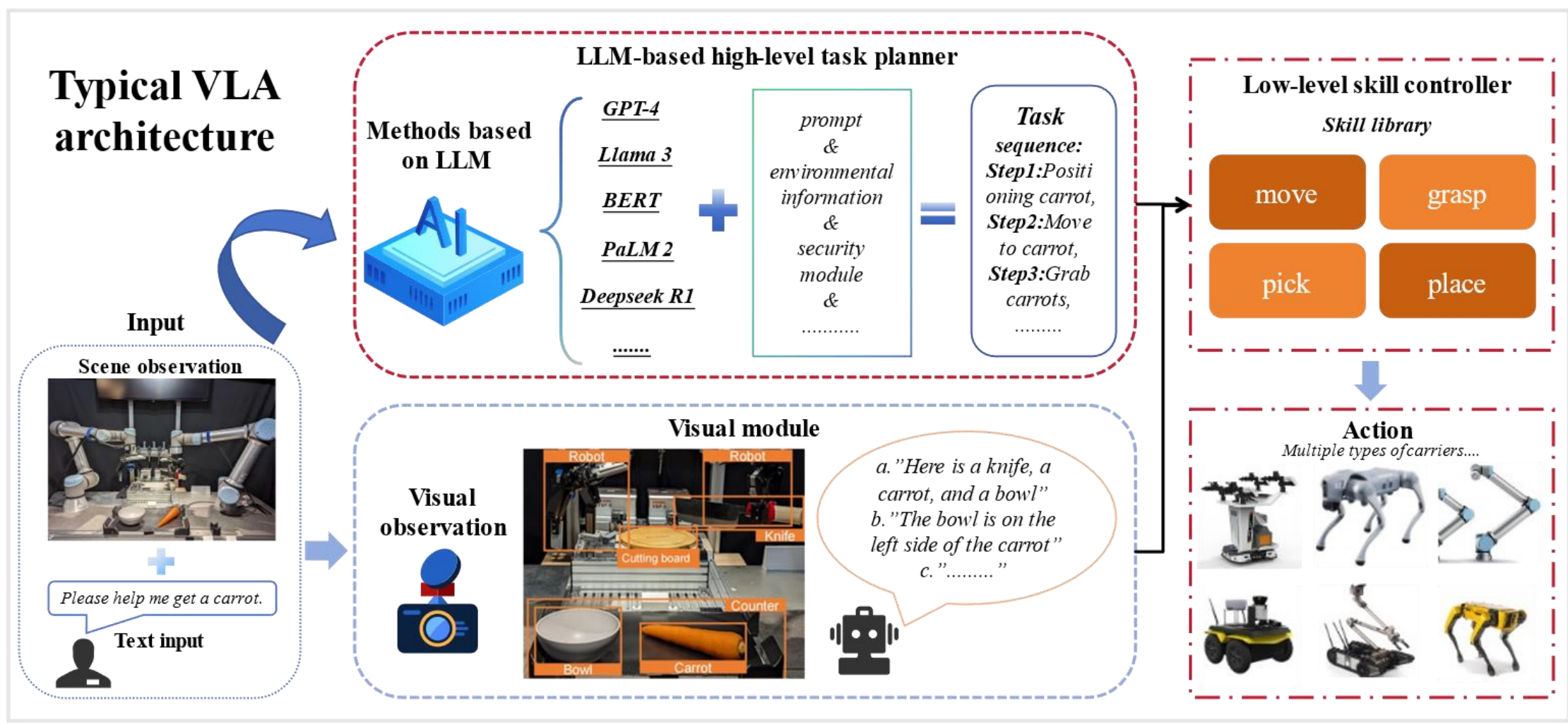

**Fig. 6.** Typical VLA architecture.

In recent years, researchers have proposed a variety of general VLA architectures, such as RT-1 [3], RT-2 [26], and OpenVLA [27], and they have continued to improve upon these architectures. To enhance the performance of the VLA model, 3D perception and reasoning was integrated into the 3D-VLA architecture [84]. This integration improved the operation capabilities of the VLA model in complex environments. A dual-process VLA framework, referred to as Dual Process VLA, that separates the complex reasoning process from the real-time motion control was developed [85]. This separation improved the operational efficiency and accuracy of the robot. The SpatialVLA [86] enhances the VLA model understanding of 3D space by the introduction of a self-centered 3D position-encoding module and an adaptive action network. Spatial Forcing [87] further improves VLA spatial awareness by aligning intermediate visual embeddings with geometric representations from pretrained 3D foundation models, thereby enhancing action precision without relying on explicit 3D inputs. The VLA model has been applied to various types of robots. For example, a flexible VLA-based operating system for dual-arm robots, referred to as Bi-VLA, was proposed [88]. This system can interpret complex human commands and perform dual-arm operations. The RoboNurse-VLA [89] applied VLA techniques to surgical-nurse robot systems. It can process surgeons' commands in real time and can accurately grasp and transfer surgical instruments. VLABench [90], a large-scale dataset and evaluation benchmark, was developed for VLA testing. It includes 100 task categories and 2,000 3D objects. This benchmark can assess the capabilities of the VLA model with respect to various tasks, particularly for long-term reasoning and multi-step planning operations.

The LLM and the VLA model have both demonstrated strong general task-planning capabilities; however, they are still primarily applied in household and open-world daily life scenarios. Due to insufficient industrial knowledge and weak industrial data perception, their direct transfer to industrial settings would not yield solutions that satisfy industrial specifications and process constraints.

## 4.2 Industrial task planning

Unlike general task planning, industrial task planning addresses production tasks in industrial settings with strict requirements and constraints. It demands higher planning accuracies, allows little room for errors, and has more serious consequences when planning mistakes occur. For example, placing a fork on either side of a bowl is acceptable for a household task, but in an industrial scenario, part A and B on a production line must stay in fixed positions, and the processing steps must follow a strict sequence. Generating the appropriate subtask sequences that meet industrial standards and constraints, is the main challenge that must be addressed when general task-planning methods are applied to industrial scenarios. This section discusses **knowledge and skill-based, learning-based,** and **LLM-based methods**.

### 4.2.1 Knowledge and skill-based methods

Most knowledge and skill-based methods involve a world model (see Section 3) and a skill library (see Section 5). These methods utilize the knowledge in the world model to reason, select, and combine skills from the skill library and thereby generate the required planning scheme. These methods can be categorized as either knowledge graph-based task planning or domain specific language (DSL)-based task planning according to the reasoning approach, as shown in **Fig. 7**.

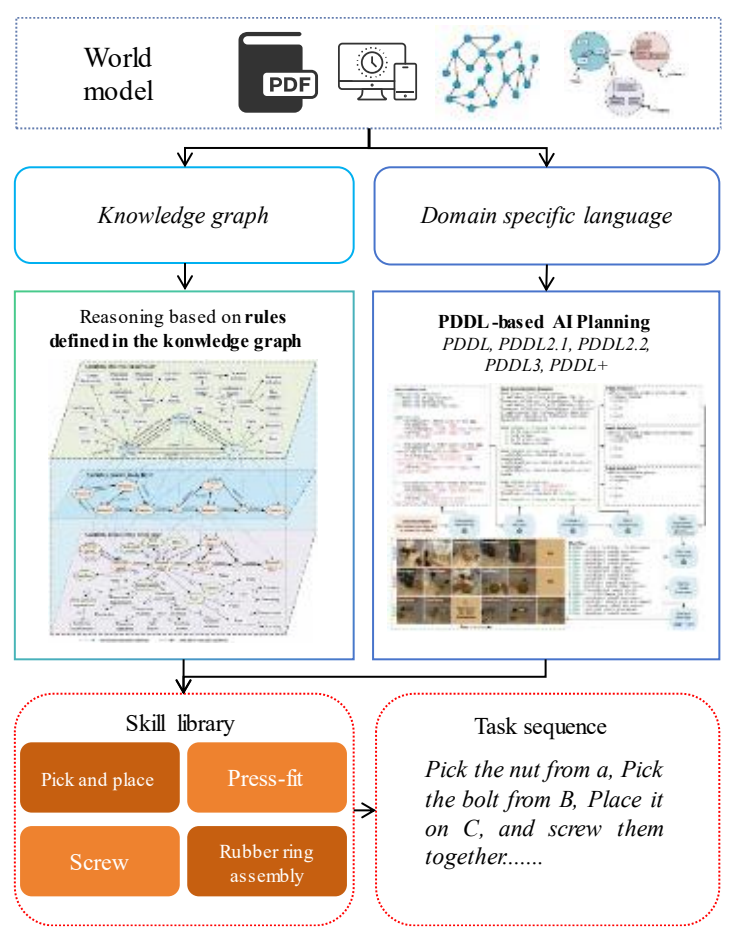

**Fig. 7.** Knowledge and skill-based task planning.

1) **Knowledge graph-based methods:** Knowledge graph-based task planning utilizes knowledge from a knowledge graph and predefined rules. Before the knowledge graph was proposed, expert systems were used for industrial task planning [91-93]. However, this method has some problems. It has fixed reasoning rules, so its generalization capability is weak and knowledge updates must be performed manually. These issues limited its application scope and led to its gradual elimination. Knowledge graphs have strong knowledge-storage capabilities. They enable convenient knowledge updates and have good reasoning capabilities. Because of these advantages, knowledge graphs have been increasingly used in manufacturing scenarios. **Table 5** lists some publications regarding knowledge and skill-based industrial task planning.

**Table 5** Surveyed literature regarding knowledge and skill-based industrial task planning.

| Method | Publication | Application scenario | Core principle | Common ground |
|---|---|---|---|---|
| Knowledge graph-based | M. Merdan et al. [94] | Industrial robot | Reasoning based on rules defined by the knowledge graph | Based on a world model, knowledge is obtained from a knowledge base for inference. Skills are obtained from a skill base and are combined. |
| | Y. Jiang et al. [95] | Automatic assembly | | |
| | T.Hoebert et al. [96] | Industrial robot | | |
| | B. Zhou et al. [97] | Automatic assembly | | |
| | Z. Qin, Y.Lu [98] | Adaptive manufacturing control | | |
| DSL-based | Z. Kootbally et al. [99] | Automatic assembly | AI planning based on PDDL | |
| | M. Cashmore et al. [100] | ROS | | |
| | F.Rovida et al. [101] | Industrial robot | | |
| | L. Heuss et al. [102] | Industrial robot | | |
| | A. Rogalla [103] | Process planning | | |

To enhance the flexibility and applicability of industrial robots, Merdan et al. [94] and Hoebrechts et al. [96] applied an ontology-based knowledge-driven framework. They proposed a new scheme to address the complex programming and high configuration costs of traditional industrial robots. To improve assembly process planning, Jiang et al. [95] explored the concept of combining digital twins with knowledge graph techniques. This approach can effectively manage complex assembly-process knowledge. Zhou et al. [97] proposed a knowledge graph-driven method of generating and evaluating assembly processes. They constructed an assembly-process knowledge graph (APKG) for the generation of assembly plans. Using interference detection and quality assessment methods, they identified feasible assembly sequences. This method was validated against an aeroengine-compressor rotor assembly case. Qin et al. [98] addressed adaptive control in large-scale personalized manufacturing by proposing a knowledge graph-based semantic representation method for dynamic manufacturing environments. By integrating factual data with machine preference information, they developed a new adaptive manufacturing control scheme.

2) **DSL-based methods:** DSL-based task planning uses specific languages, such as PDDL [104], to express and solve planning problems. PDDL, which is a standardized language for robotics planning problems, can flexibly manage complex planning issues. In recent years, it has gained popularity for industrial task-planning applications. Kootbally et al. [99] proposed a knowledge-driven method that uses a combination of knowledge and PDDL to directly convert web ontology language (OWL) to PDDL. The value that this method brings to assembly applications was discussed. The ROSPLAN [100] framework integrates task planning into an ROS. By using modular design for the knowledge base and planning system, it can automatically process planning and schedule low-level controller activities. The SkiROS [101] platform addresses the knowledge representation and autonomous task planning of robot development by employing modular design and knowledge integration. The REpac [102] framework possesses an extensible, skill-based software architecture that supports flexible configuration and autonomous task planning for industrial robots. By reusing skills and modular components, it gradually expands the reasoning capabilities of the robots so they can achieve multi-task planning. Rogalla et al. [103] proposed a domain-modeling method for discrete manufacturing that models manufacturing systems and orders in PDDL, thereby helping the planners to understand and solve problems.

Despite their excellent performance in industrial applications, knowledge and skill-based task-planning methods face challenges. These include knowledge-updating difficulties, the need for predefined rules and reasoning methods, and limited scalability for new tasks and scenarios. These issues lead to high dependencies on humans.

#### 4.2.2 Learning-based methods

Learning-based methods utilize various techniques, such as deep learning and reinforcement learning, to extract task-planning information from vast datasets. As deep learning has evolved, these methods have been applied to intelligent manufacturing scenarios [105]. They exhibit extensive application potential for many types of operations, such as robotic grasping, assembly and disassembly, process control, and industrial human–machine collaboration [106]. Learning-based methods eliminate many of the manual processes that are required by knowledge and skill-based methods. Different learning strategies have distinct advantages and disadvantages, which lead to varied application contexts and methodologies. This section summarizes three primary learning strategies, which are based on deep learning, reinforcement learning, and imitation learning, respectively, as shown in **Table 6**.

**Table 6** Surveyed literature regarding learning-based industrial task-planning methods.

| Method | Application scenario | Publications | Algorithm | Advantage | Disadvantage |
| --- | --- | --- | --- | --- | --- |
| Deep learning | Human–robot collaboration | H. Zhang et al. [107] | CNN | Processes complex data to enhance reliability | Requires a large amount of data |
| | | H. Liu et al. [108] | CNN | | |
| | Object recognition | X. Chen, J. Guhl [109] | RCNN | | |
| | | E. Solowjow et al. [110] | Dex-Net | | |
| | Subassembly recognition | C. Zhang et al. [111] | GCN | | |
| Reinforcement learning | Robot assembly | Y. Liu et al. [112] | DQN | Learning is achieved through | Unstable training may |
| | | J. Li et al. [113] | DDPG | | |
| | Robot additive manufacturing | Y. Xiong et al. [114] | DQN | autonomous interaction with the environment and policy functions | lead to unstable model performance |
| | Logistics robot | F. Fan et al. [115] | DODPG | | |
| | Automatic assembly | M. Jiang et al. [116] | DRL | | |
| Imitation learning | Robot assembly | Y. Wang et al. [117] | GMM+GMR | Extensive mathematical modeling and optimization is not needed. | Requires a large amount of high-quality sample data for training |
| | | S. Scherzinger [118] | LSTM | | |
| | | T. Zhang [119] | p-LSTM | | |
| | | S. Ji et al. [120] | DDPG | | |

1) **Deep learning-based methods:** Deep learning-based methods can process complex production data and easily uncover hidden data patterns. However, their heavy reliance upon data limits their use in data-scarce scenarios. Using CNN and LSTM techniques, Zhang et al. [107], predicted human assembly actions. Liu et al. [108] developed a CNN-based multi-modal user interface that enables non-professionals to easily control robots. Chen et al. [109] used the RCNN algorithm to achieve object recognition in work areas during industrial robotic grasping processes. Solowjow et al. [110] created a DEX-Net-based grasping robot that exhibited a high success rate. Zhang et al. [111] used model-based design (MBD) to integrate geometric and engineering information, then they constructed heterogeneous knowledge graphs and used the GCN algorithm to identify subassemblies.

2) **Reinforcement learning-based methods**: Reinforcement learning-based methods excel in scenarios that have limited data and that require independent decision-making because they can learn autonomously through interactions with the environment. However, the randomness of the training process can produce unstable results. Liu et al. [112] presented a vision-combined reinforcement learning scheme to aid human–machine interactions during manufacturing. It allows robots to observe human-collaborator information and then to adjust their decisions and actions. To address multi-variety and small-batch assembly issues, Li et al. [113] combined deep reinforcement learning and a digital twin. They established digital-twin models and trained reinforcement learning models to plan assembly processes and predict production line faults. Xiong et al. [114] applied the Kriging dynamic function to additive manufacturing scenarios. This method enables learning through multiple agents and workspaces, and thereby reduces material consumption during additive manufacturing. To improve assembly-sequence planning for aviation products, Jiang et al. [116] developed a novel fine-grained assembly-sequence planning method by combining a knowledge graph and deep reinforcement learning. To address the internal logistics of the manufacturing industry, particularly in complex workshop environments, Fan et al. [115] proposed a navigation method that utilizes deep reinforcement learning and wheeled mobile robots. By the adoption of a dynamic-observation Markov decision process and distributed scene training, this method achieved efficient scene modeling and path tracking control in complex industrial settings.

3) **Imitation learning-based methods:** Imitation learning-based methods enable faster learning than reinforcement learning-based methods and require less mathematical modeling and optimization. They are ideal for scenarios with clear tasks and abundant human expert demonstration data. The results of these methods are more deterministic than those of the other methods; however, they require many high-quality demonstrations. In recent years, imitation learning-based methods have been widely used in robot-assembly scenarios [117-120]. These methods utilize demonstrations as learning data for the robots, thereby enhancing the assembly task-planning capabilities of the robots.

Learning-based methods eliminate the need for the manual rule definitions that are required by knowledge and skill-based methods; however, they require substantial data support and demand significant computational power and training time. The models are typically trained for specific scenarios or tasks, so they have only average generalization capabilities; thus, it is difficult to transfer them to other tasks.

#### 4.2.3 LLM-based methods

During industrial task-planning operations, LLM-based methods leverage the powerful text generation and understanding capabilities of LLMs. They can process complex industrial documents, operation manuals, and user feedback, thereby enhancing the information-processing efficiency and accuracy. These methods utilize externally acquired knowledge and the reasoning capabilities of LLMs to achieve task planning. As AI technology has rapidly advanced, LLMs have significantly impacted the industrial field because of their natural language understanding and multi-modal information-processing capabilities [5, 121]. During high-level industrial-robot task-planning processes, LLMs can effectively interpret fuzzy input tasks and decompose them into a series of subtasks. Unlike other methods, LLM-based methods do not require many human-defined rules or extensive training data. Although they have been studied extensively in connection with general task-planning processes, research regarding the use of LLMs for industrial applications is still in its early stages. Most of the existing work has focused on the reasoning capabilities of LLMs, and some studies achieved positive results by fine-tuning LLMs with new data.

LLM-based industrial task-planning methods have been used for various manufacturing tasks. Tanaka et al. [122] used LLMs to develop a voice-controlled control system for a polishing robot. They analyzed natural language using the GPT-3 model and converted it into numerical commands, which enables users to control robot actions through voice inputs. This approach allows workers to use robots for specific functions without the need for complex programming. Wang et al. [123] proposed an LLM-based visual language navigation method for intelligent manufacturing systems. This method involves three steps: the reconstruction of real-world manufacturing scenes using 3D point clouds, the instigation of navigation actions with the LLM code-generation capability and path planning with the Pathfinder algorithm, and the generation of executable robot actions. Fakih et al. [124] used LLMs to achieve verifiable PLC programming in industrial control systems and introduced the LLM4PLC framework. By using engineering prompts and low-rank adaptation (LoRA) to fine-tune the model, and by incorporating user feedback and external tools to guide the LLM generation process, they verified the system using the Fischertechnik manufacturing test platform (MFTB). This method significantly reduces the time required to write the PLC code and improves the quality of the LLM-generated PLC code. Fan et al. [11] explored the application of LLMs to industrial robots and proposed a framework for the independent design, decision-making, and task execution of industrial robots. This framework uses LLMs to extract the manufacturing tasks and process parameters from natural language, select end effectors, generate motion paths according to predefined conditions, and evaluate the path effectiveness. It then uses skills in the code and task bases to complete manufacturing tasks. Gan et al. [125] proposed a bionic robot controller that can satisfy the autonomous task-planning needs of the manufacturing industry. This controller combines motion control, visual perception, and autonomous planning modules to achieve multi-object rearrangement functions. Gkournelos et al. [126] applied LLMs to manufacturing systems to enhance human–machine interactions in factory settings. Their system is based on extensible components that can be classified as either agents or modules. The agents include formatting, interactive, and manufacturing agents that possess natural language processing capabilities. The modules include robot-behavior planning and human–machine interaction modules. When it was tested in inverter and industrial air compressor assembly scenarios, the system achieved positive results. Xu et al. [127] discussed the application of EI techniques to additive manufacturing processes. They studied methods of causing 3D printers to interact with the environment as organisms do by investigating biological growth processes. For fixed 3D printers, this approach can automatically generate tool paths and machine code using basic models; thus, it reduces the demand for expert knowledge.

LLM-based methods have also been used for automatic electric-vehicle disassembly, industrial drones, construction robots, and other applications. To solve electric-vehicle battery disassembly problems, Peng et al. [128] used neural symbol AI to develop an autonomous mobile-robot system for battery disassembly (BEAM-1). The system used neural predicates and action primitives to achieve environmental perception and autonomous planning. It also employed an LLM heuristic search in the planning process, thereby improving efficiency and addressing search-space explosion issues. Zhao et al. [129] proposed an AeroAgent architecture for industrial drones. It treats agents as the brain and controllers as the cerebellum in industrial tasks. The MLLM-based agents can analyze multi-modal data, customize plans based on environmental information, and quickly adapt to new tasks using small-sample learning. The ROSchain framework integrates MLLMs with an ROS, thereby enabling direct control of the drone actions and ensuring that the input matches the actuator capabilities. You et al. [130] applied EI techniques to construction robots and proposed the Dexbot framework. This framework contains six key steps, by which it achieves robot flexibility and adaptability for three primary construction tasks: structural assembly, material processing, and quality inspection.

Although LLMs have been applied in some industrial scenarios, most of these applications only use LLMs as auxiliary components in the task-planning process. The core methods by which industrial-robot task planning is achieved are still based on traditional techniques. The LLMs are primarily used for optimization and improvement of traditional methods rather than for complete replacement or redefinition of these methods.

To summarize Section 4.2, existing industrial task-planning methods can be divided into three main categories: knowledge and skill-based methods, learning-based methods, and LLM-based methods. The knowledge and skill-based methods rely too heavily upon artificially determined rules. When new scenarios or tasks are encountered, professionals must often redefine the relevant rules; thus, these methods have low flexibility. Learning-based methods require large amounts of data and computing resources for training, and they must often be retrained when for new scenarios or tasks; thus, they possess limited generalization capabilities. It is difficult for methods of these two types to effectively meet the small-batch and customized-manufacturing requirements. However, due to limitations in training resources, the emerging LLM-based methods often fail to fully understand the professional knowledge in industrial settings; thus, their effects are still mediocre when they are used for industrial task planning.

A potential solution is proposed for the problems discussed above: a combination of an LLM with the retrieval-augmented generation (RAG) technique. By establishing a specific external knowledge base for use as the world model, the LLM can master the knowledge unique to the industrial field and then improve its ability to answer industrial questions. At present, a few studies regarding such a method have achieved some results; however, this method has not yet been directly applied to industrial task planning. Bei et al. [131] proposed a question-and-answer system that is based on the integrated term enhancement method. By accurately extracting and interpreting key terms from knowledge documents and building a term dictionary, it can enhance the query capability. The AMGPT [132] question-and-answer system is based on a combination of the pre-trained LLaMa 2-7B model and RAG; it enhances the question-answering capability in additive manufacturing scenarios by the dynamic integration of information. Other studies utilized the LLM + RAG method to process a large amount of data during industrial-production processes [133-135] with the goal of maximizing the use of these data for prediction and decision-making tasks. In other research, an LLM was combined with a knowledge graph in an industrial setting [136, 137]; the generalization capability of the large model was used in each task and the accurate reasoning rules in the knowledge graph were used to improve the performance of the LLM for specific industrial tasks.

The goal of this scheme is to fully utilize the powerful representation learning capabilities of LLMs and the advantages of the RAG technique for knowledge integration and retrieval. This method can enhance the generalization capability and adaptability of industrial task planning while improving its accuracy. The application of this scheme to industrial-robot task planning can mitigate the shortcomings inherent to existing methods, thereby better meeting the flexible-production and customized-manufacturing needs of the manufacturing industry.

# 5 Low-level skill controller

The low-level skill controller, which serves as the core hub that connects the high-level task abstraction with the physical execution of the robot, is used to convert decomposed subtasks into a series of skills and to output executable programs. This paper does not emphasize the skills discussed in the previous EIR reviews [2, 138] (such as perception, navigation, manipulation, and other skills that are used in unstructured environments); rather, it focuses on the unique skill paradigms required by industrial scenarios, and especially those used in assembly tasks. The low-level skill controller utilized by EIIR possesses a two-layer "skill-language" architecture, in which the skill layer uses modular encapsulation to convert actions into reusable skills. The language layer uses DSL and constraint rules to achieve physical execution of the skills. The transformation from subtasks to skill sequences primarily relies upon two approaches: knowledge and skill-based methods and LLM-based methods. Both of these approaches were discussed in Section 4. This section presents analyses of the work related to the two layers; therefore, it provides a reference for readers who wish to select appropriate skill libraries and control languages for the formation of low-level skill controllers.

## 5.1 Industrial skills

Industrial skills are the standardized, reusable, and programmable units provided by robots or other equipment to achieve specific manufacturing goals (such as assembly, welding, and inspection) in a structured or semi-structured industrial environment. By means of hardware–software encapsulation, these skills abstract the underlying sensor data, control algorithms, and actuator actions into process semantics-oriented functional interfaces; thus, they transform complex physical interactions into programmable industrial behavior modules. Along with the rigid constraints and task requirements of industrial scenarios, the "*Task–Subtask–Skill–Action*" concept levels are proposed here, and standardized mapping of complex processes is achieved by means of level-by-level decoupling. Using a shuttle-valve assembly task as an example, **Fig. 8** introduces the relationships between the four concept levels. Details regarding each level are also provided in the following list:

1) *Task*: A high-level production activity with a complete functional objective, such as "assemble shuttle-valve" or "weld PCB board", that encapsulates the process semantics. As stated in Section 4, an EIIR can decompose tasks into subtasks using a high-level planner that combines an LLM with RAG and that is guided by process knowledge from an industrial knowledge graph.

2) *Subtask*: A fundamental process unit that is analogous to a "procedure" in an MES. For example, the "assemble shuttle-valve" task may be split into four subtasks: "assemble large rubber ring", "place steel balls", "assemble small rubber ring", and "press piston". subtasks are semantically linked to skills in the knowledge graph, which also transforms their process requirements into execution parameters.

3) *Skill*: An abstracted capability that is provided by one or more devices and is responsible for transforming subtasks into executable physical actions. Skills have two key attributes: cross-device collaboration (e.g., a "press-fit" skill may coordinate the actions of multiple devices, such as robot-arm motion, force-sensor feedback contact state, and vision-system correction orientation) and a logical container (which defines action sequences using finite state machines (FSMs) or temporal logic).

4) *Action*: An atomic physical operation that is performed by a device. Actions are often directly linked to hardware via DSLs or protocols (e.g., EtherCAT and PROFINET), and they lack context awareness. For instance, a "grasp" action is initiated by a Boolean signal, but it cannot detect object presence. Actions standardize device interfaces by decoupling vendor-specific APIs from skills. For example, differing joint controls of various manipulator brands are abstracted behind a "move" action that exposes only generic parameters, such as the orientation, speed, and acceleration of the target.

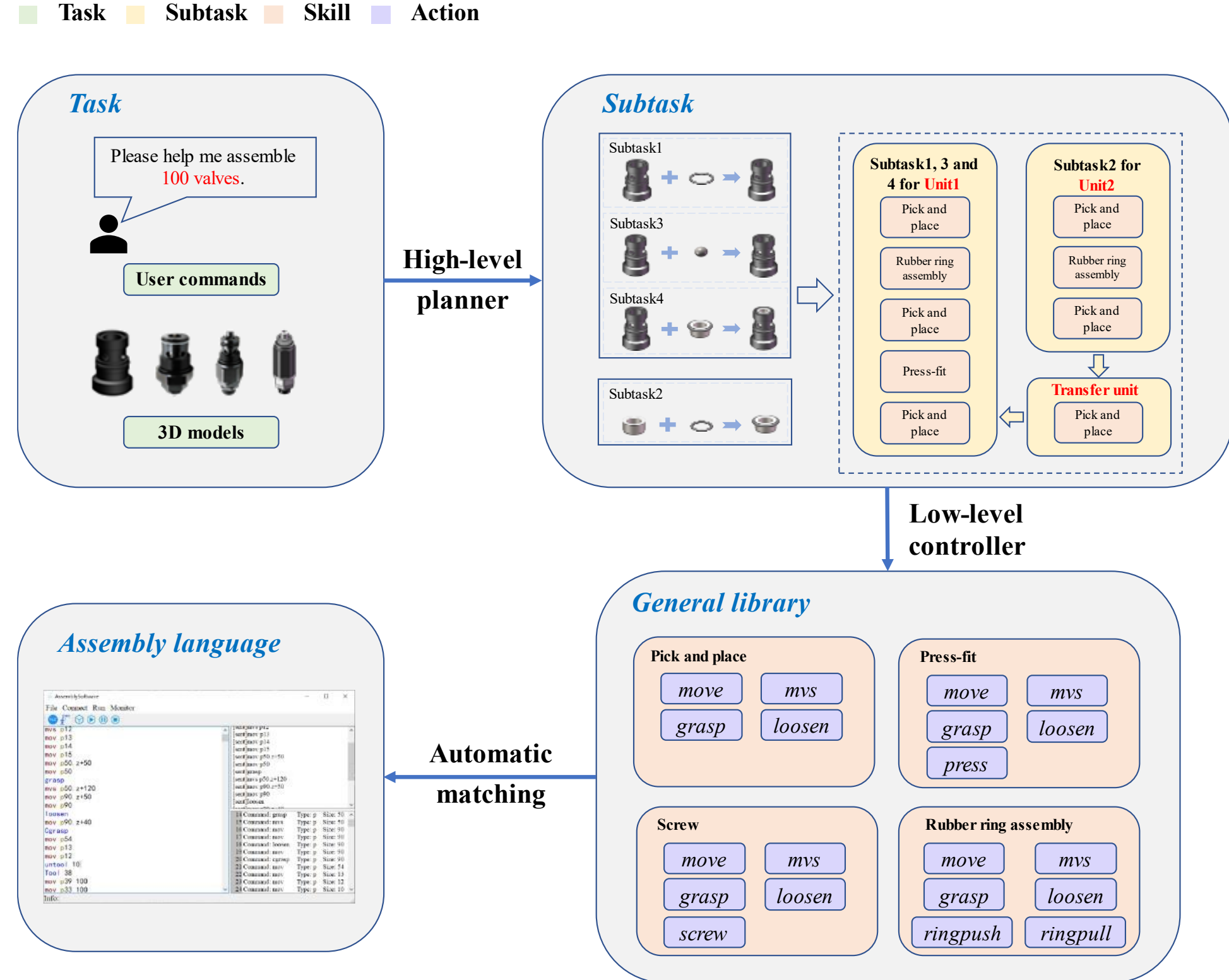

**Fig. 8.** Example of the "*Task–Subtask–Skill–Action*" architecture. The task is usually a description based on user-command input and 3D models. After it is decomposed by the high-level task planner, the shuttle-valve assembly task is split into a series of subtasks, which are assigned to different units for execution. Each subtask is automatically matched with skills from the general library by the low-level controller. Finally, the overall task is transformed into a program that consists of a combination of the lowest-level actions.

In the architecture described above, the construction of standardized skill libraries and action libraries is fundamental to the execution of industrial tasks. These libraries provide reusable and extensible foundational capabilities that enable the rapid composition of complex processes by means of modular encapsulation. Reinhart et al. [139, 140] reviewed a range of publications, standards, and studies in the skill taxonomy and ontology field, after which they proposed a skill-classification method specifically for assembly. Lee et al. [141] identified nine atomic actions that are commonly used in assembly processes. Building upon these classification schemes, the authors of the current work conducted a literature survey to analyze and categorize recent developments. **Fig. 9** defines 15 commonly used robotic actions, the taxonomies of which follow two principles:

- *General and Specific Actions*: General actions, which range from "**Move**" to "**Sense**," include the majority of common industrial operations and can typically be executed using only a robotic arm and simple tools. In contrast, specific actions require special tools or additional material support. For instance, the "**RingPush**" action involves the insertion of an elastic seal ring into a groove by means of precise axial pressing that is performed by a dedicated end-effector, while the "**Print**" action requires an integrated material feeding system and a dispensing nozzle. Beyond the eight specific actions listed, users may extend the library according to the application needs (e.g., “drill” and “mill” actions can be added for subtractive manufacturing applications).

- *Semantic Merging*: The taxonomy consolidates actions with similar semantics that are found throughout the published literature. Translational actions, such as “Push,” “Retract,” and “Slide” are abstracted into a parameterized variant, “**Move**.” Similarly, assembly actions, such as “Insert,” “Snap in,” and “Mount” are unified under the “**Press**” label, which encapsulates force-controlled pressing behaviors. This merging strategy streamlines the action library for more efficient reuse and system integration.

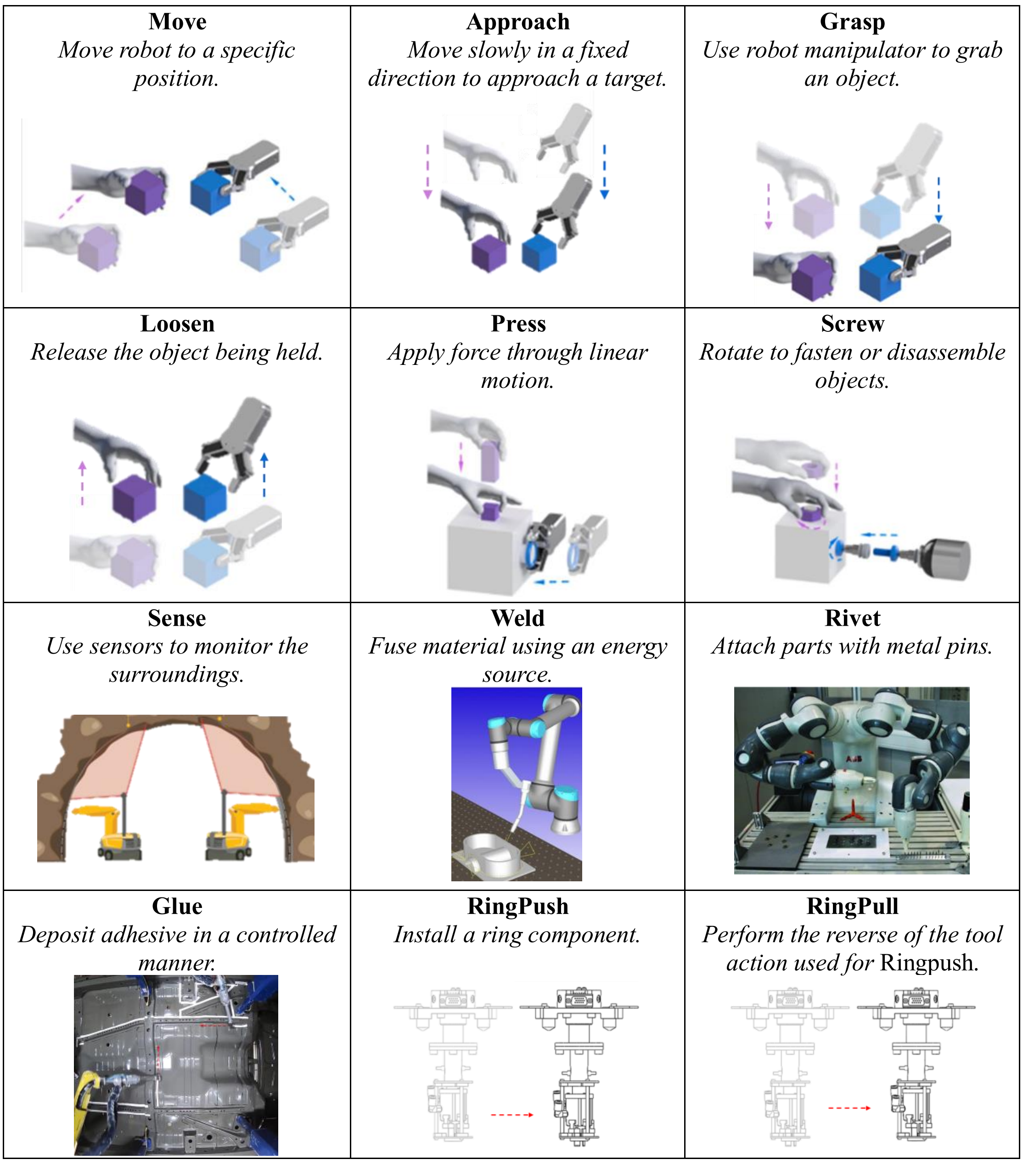

| **Move**<br>*Move robot to a specific position.* | **Approach**<br>*Move slowly in a fixed direction to approach a target.* | **Grasp**<br>*Use robot manipulator to grab an object.* |
| --- | --- | --- |
| **Loosen**<br>*Release the object being held.* | **Press**<br>*Apply force through linear motion.* | **Screw**<br>*Rotate to fasten or disassemble objects.* |
| **Sense**<br>*Use sensors to monitor the surroundings.* | **Weld**<br>*Fuse material using an energy source.* | **Rivet**<br>*Attach parts with metal pins.* |
| **Glue**<br>*Deposit adhesive in a controlled manner.* | **RingPush**<br>*Install a ring component.* | **RingPull**<br>*Perform the reverse of the tool action used for* Ringpush. |

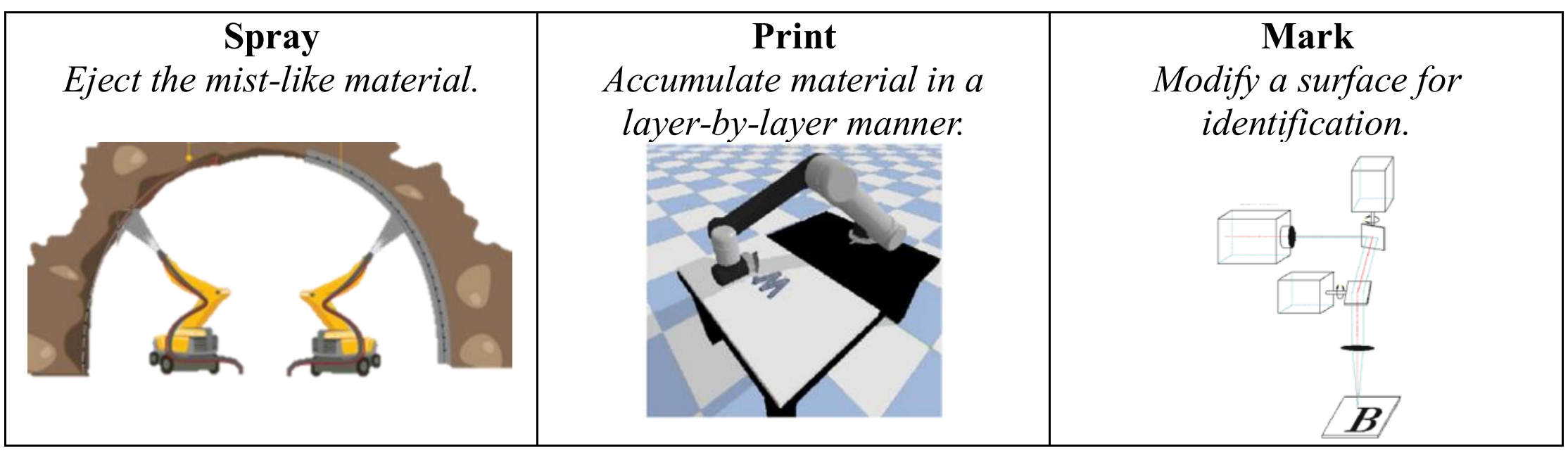

**Fig. 9.** Robotic action taxonomy from the literature review. The illustrations were adapted from those of previous publications [11, 141-145].

In the research and application of robotic skills, the skills can be functionally divided into four main categories: handling, joining, inspecting, and special operations. These four skill types are summarized in **Table 7**. Each type consists of multiple underlying actions, and various skills can be composed into subtasks that enable robots to execute complex industrial tasks. The handling skills, such as "Pick and place" and "Transport," form the foundation of robotic manipulation and are extensively applied in automated production, warehousing, logistics, and service robotics operations. Notably, "Pick and place" was featured in 80% of the surveyed literature; its presence is essential in any skill library. "Transport" involves robot navigation and material transfer, and it is often combined with environmental perception that is enabled by sensing actions so that it can be performed effectively.

The joining skills primarily involve the connection of components, and they typically require high-precision orientation alignments, force-control strategies, and sensor feedback. These skills are fundamental to industrial robots, and particularly to electronic assembly, mechanical manufacturing, and other related applications. "Press-fit" and "Screw" are the joining skills that are most frequently referenced in the literature; this frequency reflects their broad applicability. In contrast, "Weld," "Rivet," and "Glue" are more specialized skills and are commonly employed in certain sectors, such as automotive manufacturing and metal processing. Notably, while "Ring assembly" is rarely discussed by existing publications, it is a prevalent skill in real-world production settings and has therefore been included in the construction of the skill base.

The inspecting skills include both the "Measure" and "Check" skills. Although both of these skills involve detection, they have different focuses. "Measure" focuses on acquiring quantitative data, such as size, position, or force information. It uses various tools, such as force/torque (F/T) sensors and visual sensors, to accomplish this goal. These measurements provide critical input or output parameters for the handling and joining skills. In contrast, "Check" is oriented toward the detection of anomalies. It assesses whether an action has been successfully completed or if the state of a system deviates from predefined thresholds. It is typically implemented through tactile sensing, collision detection, or force feedback-based anomaly analysis. These inspecting skills are crucial to quality assurance, automated testing, and predictive maintenance in industrial applications.

Special operations skills include some other specific tasks that are performed by robots, such as "Spray," "3D-print" and "Mark." These skills are usually specific to certain manufacturing or processing requirements. "Spray" and "3D-print" are related to coating spraying and additive manufacturing, respectively. "Mark" is very common in real production lines; it includes laser marking and ink-jet marking, among other applications, and is worth preserving in the skill library.

According to the frequency with which they were referenced, certain skills were determined to be indispensable to the construction of an industrial-robot skill library. These skills include “Pick and place,” “Transport,” “Press-fit,” “Screw,” “Measure,” and “Check.” These skills cover the essential grasping, motion, assembly, and inspection capabilities; thus, they form a foundational skill set for virtually all industrial and service robot systems. Notably, several articles [142, 146-148] present frameworks that incorporate more than four skills and propose comprehensive robot-control architectures that are centered around skill modularization. These references offer valuable insights and are particularly recommended for readers that wish to design robust and extensible skill libraries.

**Table 7** Taxonomy of robotic skills and 75 related articles. Some of the references were obtained from the survey conducted by Pantano et al. [149], while the rest of the publications were identified through a systematic Web of Science search. The search strategy employed the Boolean query: ‘*robot AND skill AND (industry OR manufacturing)*’. All the retrieved articles were rigorously reclassified under the proposed skill scheme.

| Skill category | Skill | Included actions | References | Parameters |
|---|---|---|---|---|
| **Handling** | Pick and place | move, approach, grasp, loosen | [65, 101, 102, 120, 144, 146-148, 150-202] | End_pose, Vel, Acc, Move_type (PTP/LIN/CIRC) |
| | Transport | move, sense | [101, 140, 142, 147, 148, 150, 155, 167, 172, 173, 176-178, 180, 182, 184, 186, 189, 196, 203-206] | Path_plan (RRT/PRM), Vel, Payload_mass, Env_map, Obst_avoid |
| **Joining** | Press-fit | move, approach, press | [65, 120, 146-148, 150, 151, 156, 157, 160, 161, 163, 166, 168, 170, 171, 179, 181, 183, 186-188, 193, 202, 207-210] | Force_set, Insert_depth, Align_tolerance |
| | Screw | move, approach, screw | [140, 146, 147, 153, 164, 165, 171, 189-191, 209, 211, 212] | Torque_set, RPM, Screw_type (M3/M4), Ang_alignment |
| | Weld | move, approach, weld | [140, 143, 171, 208] | Voltage, Current, Feed_ speed, Weld_path, Gas_flow |
| | Rivet | move, approach, rivet | [144, 150] | Impact_force, Align_offset |
| | Glue | move, approach, glue | [150] | Dispense_vol, Glue_path |
| | Ring assembly | move, approach, grasp, loosen, ringpush, ringpull | - | Spread_force, Ring_dim (ID/OD), Align_guide |
| **Inspecting** | Measure | move, sense | [101, 120, 146, 150, 154, 155, 159, 161-163, 167-170, 172, 173, 175-178, 180, 181, 197-206, 208, 211, 213, 214] | Sensor_type (Laser/Force), Accuracy, Data_output |
| | Check | move, sense | [102, 140, 146, 158-160, 163, 175-178, 181, 186, 194, 196, 207] | Sensor_type (Laser/Force), Check_value, Tolerance |
| **Special operations** | Spray | move, approach, spray | [142, 204] | Paint, Flow_rate, Nozzle_vel |
| | 3D-print | move, approach, print | [142] | Layer_height, Print_speed, Nozzle_temp |
| | Mark | move, approach, mark | - | Depth_set, Mark_speed |

In addition to the skill library, the parameter settings affect both the task success and the overall efficiency. **Table 7** lists the parameters that were most commonly used in the literature. From the perspective of motion control, velocity (Vel) and acceleration (Acc) are the core parameters of all the motion-related skills, determining the stability and response speed of the action. Path-planning parameters, such as the rapidly-exploring random tree (RRT) and the probabilistic road map (PRM), primarily affect the autonomy and environmental adaptability of the robots by means of the algorithm, and they particularly affect the robot obstacle-avoidance capability in complex or dynamic environments. Force control parameters (Force_set and Torque_set) are very important for skills that involve physical interactions, such as “Press-fit” and “Screw.” If the values are too large, the workpiece will be damaged; however, values that are too small may lead to assembly failure. In addition, tolerance parameters (Align_tolerance and Check_value) reflect the skill-errors tolerance of the users; they are also closely related to the sensor accuracy. For some special material-processing skills, such as “Weld,” “Spray,” and “3D-print,” the temperature, flowrate, and feed rate directly determine the uniformity and final quality of the material deposition. It is worth noting that some parameters affect not only a single skill, but also the stability of the whole task. For example, the Payload_mass of a robot affects the motion-trajectory planning, while semantic map of the environment (Env_map) affects the long-term path optimization. Therefore, when constructing a skill library, the selection of the skill parameters must not only be optimized for a single execution, but must also consider the parameter interactions at the global level to ensure that the robot can effectively perform skills in a variety of application scenarios.

In EI research, skill libraries are central role to the low-level skill controller because they serve as structured interfaces between the LLMs and the physical environment. The PRoC3S [215] uses an LLM to generate parameterized skill codes (which include actions, such as grasping or placing, and coordinate parameters). These skill codes are combined with a continuous constraint satisfaction problem (CCSP) solver to process kinematic, geometric, and physical constraints. This strategy enables natural language-driven drawing and stacking tasks. The PromptBook [216] is an instruction–example prompt method. It describes skill parameters (such as the orientation coordinate system) and physical constraints (such as the robot-arm accessibility) using application programming interface (API) documents, thereby allowing the LLM to generate new skill codes, such as drawer switches, with zero samples. Wiemann et al. [217] encapsulated skills as the services of ROS2 and used an LLM to analyze the implied semantics in natural language (for example, it maps the command "move the camera 10 mm" to a moving relative coordinate service and automatically fills in the parameters). This approach increases the ease of user programming. The LiP-LLM system [218] was designed to build a skill-dependency graph and extract temporal logic from language text, such as "Clear A before placing B." It then uses linear programming to optimize task allocation among multiple robots.

These methods, which all involve combining LLMs with skills, reveal that the code-generation paradigms that are based on structured skill bases have become the mainstream implementation mechanism for LLM-driven EI. In this method, a strong encapsulation interface is generally used during the design of the skill library so that the skill code generated by the LLM can be directly mapped to the physical execution. This design paradigm is primarily reliant upon two types of structured languages. The first type is robot middleware interface languages (such as the ROS2 [217]), which achieve skill–device matching through predefined service types and message structures. The second type is domain-specific programming interfaces (such as the robot API description [216]), which ensure that the generated code is physically enforceable by strictly restricting the function signature and coordinate system. For industrial control, however, such interfaces must be connected with PLC ladder diagrams or structured text (IEC 61131-3) in addition to robots. The establishment of a unified DSL that is compatible with various types of controllers is crucial to the achievement of LLM-driven EIIR.

## 5.2 Low-level control language

In the low-level skill controller of the EIIR framework, a language is often needed to express the skills and thereby facilitate the actuator understanding. This kind of language must possess dual attributes: it must retain skill-semantics abstraction to support low-level reasoning as well as be capable of embedding hardware-interface specifications to ensure physical enforceability. DSL effectively fills the gap between skill abstraction and hardware instructions by means of a semantic layered architecture. Van deursen et al. [219] defined DSL as a programming language or executable specification language that offers, through appropriate notations and abstractions, expressive power that is focused on, and is usually restricted to, a particular problem domain. Its core feature is that it provides natural abstraction that is friendly to domain experts while maintaining strict machine processability. This feature is particularly important in industrial control scenarios. To address the problem of interface fragmentation of heterogeneous devices, such as robot controllers and PLCs, DSL can not only encapsulate the underlying languages of different control protocols, but can also inject reasoning rules by using coordinate-system constraints and kinematics rules to build a standardized semantic model. For example, an industrial DSL can define a unified "Pick and place" skill, and its parametric interface can automatically map to the RAPID command of an ABB robot and the ST of a Siemens PLC. This semantic design paradigm causes DSL to be the preferred technology carrier for cross-controller code generation and runtime verification.

In 2015, Nordmann et al. [220] systematically sorted out the robotic DSLs and divided them into nine categories according to Part A of the Springer Handbook of Robotics [221]. However, this taxonomy primarily focused on general robotics, and it is difficult to adapt it to the special requirements of heterogeneous equipment and executability in industrial control scenarios. During the current study, DSL literature from the industrial robotics field was retrieved for the years 2016–2025. The engineering value of the DSLs was evaluated using seven metrics, as shown in **Table 8**:

1) *Kinematics*: Kinematics are used to determine whether a DSL has the basic ability to drive robot motion. A robot API (such as MoveIt! or an ROS) can be called or joint-control instructions can be directly generated to verify the completeness of the DSL kinematic modeling.

2) *Path planning*: Path *planning* is used to evaluate the flexibility and configurability of the trajectory planning. The minimum standard is the support of basic navigation-path generation, while the high-order requirements include speed or acceleration curve parameterization and dynamic obstacle avoidance.

3) *Real-time*: This metric indicates whether the control instructions are online schedulable. This is different from the code-import mode that is used after offline programming because hard real-time features, such as runtime task interruption (e.g., emergency stop) and priority preemption, must be supported.

4) *Perception–Action*: This metric is used to verify the dynamic correction capability of the perception data with respect to execution logic. Typical implementations include sensor-event triggering-state migrations, such as visual positioning-error triggering relocation. Conditional branching and asynchronous event-processing mechanisms must be provided at the DSL syntax level.

5) *PLC*: PLC is used to determine whether a DSL is interoperable with industrial controllers. The DSL must support the generation of IEC 61131-3 code (such as structured text), or data exchange with PLC through the OPC UA protocol.

6) *Tool chain*: This metric primarily focuses on the graphical user interface (GUI) development environment and the simulation verification capability. The former requires integrated graphical interfaces (such as debugging tools), while the latter requires seamless docking with the simulation platform to achieve control logic verification.

7) *Industrial application*: Although this is not a necessary technical indicator, an actual scenario verification can aid in the optimization of the DSL robustness during the design process. In particular, it can provide empirical feedback regarding exception handling and long-term operation stability.

**Table 8** Overview and evaluation of the 20 surveyed DSLs obtained for the years 2016–2025. The search strategy employed the Boolean query: *'(domain specific language OR domain specific modeling language OR dsl) AND (robot OR robotic) AND (industry OR manufacturing)'*. In the table, √ indicates that the DSL meets the standard, while ○ indicates that it partially satisfies the standard.

| DSL | Kinematics | Path planning | Real-time | Perception –Action | PLC | Tool chain | Industrial application |
|---|---|---|---|---|---|---|---|
| Reversible Execution [222] | | | | √ | | | Product assembly |
| Web-Application [223] | | | | | √ | ○ | Modular production plants |
| Block-based language [224] | | | | | √ | ○ | Automotive manufacturing |
| RoboticSpec [225] | √ | | √ | | | | Failure detection |
| BDD [226] | √ | | | √ | | | - |
| DSL in wood [227] | √ | | | √ | | | Wood manufacturing |
| GeometrySL [228] | √ | | | | | √ | Medical robot |
| LoTLan [229] | | √ | √ | √ | | | Warehouse logistics |
| RoboLang [230] | | √ | | √ | | ○ | Healthcare robot |
| Salty [231] | | √ | | √ | | ○ | UAV |
| PyDSLRep [232] | | √ | | √ | | √ | Mobile robot |
| CAPIRCI [233] | √ | | | √ | | ○ | Collaborative robots |
| EzSkiROS [234] | √ | √ | | √ | | | - |
| SMACHA [235] | √ | | √ | | | √ | - |
| Assembly [236] | √ | | √ | | | √ | Product assembly |
| RoboSC [237] | | √ | √ | √ | | ○ | ROS supervisor |
| PDDL [184, 238] | √ | √ | | √ | | ○ | Kitting |
| RoboArch [239] | √ | √ | √ | √ | | √ | Nuclear robotic systems |
| UMRF [240] | √ | √ | √ | √ | | ○ | Remote inspection |
| A-code [241, 242] | √ | | √ | √ | √ | ○ | Product assembly |

In an LLM-driven skill library, the DSL achieves direct mapping from natural language to physical execution by binding structured skills. Various DSLs place different emphases on PLCs, navigation path planning, and robot motion, among other metrics. For example, block-based language [224] deeply integrates blocky programming with industrial automation. It achieves seamless connections between graphical modules and Rockwell or Siemens hardware, and it can be used to construct a semantic mapping-based PLC verification system. The Lotlan approach [229], which is based on natural language processing and DSL, can achieve cooperation frameworks between humans and mobile robots. Its core innovations are the transformation of voice inputs into standardized task descriptions (subject-verb-object structure) and support of the automated guided vehicle (AGV) dynamic task scheduling through a lightweight syntax that separates logic and control. SMACHA [235] is a meta script-based DSL that is used for templating and code generation. It simplifies robot-skill arrangement through declarative YAML script, supports modular skill encapsulation (such as grabbing or placing), and achieves efficient skill combination and reuse during complex tasks. Heuss et al. [184] proposed a PDDL-based automatic planning domain adaptation method. By dynamically associating an abstract planning model with parameterized robot skills, a planning domain description that is oriented to specific assembly scenes can be automatically generated. Thus, non-professional users can achieve industrial-robot autonomous task planning by merely configuring skill parameters. Wanna et al. [240] proposed the unified meaning representation format (UMRF) and a task-planning framework for industrial scenarios. In this method, an LLM converts natural language into UMRF graphs in the JSON format. Each node corresponds to executable robot skills (such as navigation, grabbing, and scanning), and it supports sequential, concurrent, and circular structures. A-code [241, 242] was the only DSL-supporting cooperative robot–PLC control that was found during the survey. Its syntax and four-level architecture enable modular assembly programming. An GUI and cross-device synchronization cause it to be operable on reconfigurable flexible assembly lines [49, 243]. In general, these DSLs provide a flexible and scalable technical basis for industrial automation and robot control systems, and they serve as the intermediate entities that help agents to seamlessly connect with hardware. To date, the agents have transformed the natural language into executable programs for devices.

# 6 EIIR simulators

EIIR simulators are high-fidelity virtual platforms, which can simulate the motion and manipulation process of robots in real industrial environments by accurately modeling the hardware, dynamics, sensors, and control logic. They can both reproduce the physical behavior of the robots and simulate the environmental interference and action feedback; thus, they can provide data for robotic algorithm development. With the help of a simulator, a large amount of training data can be generated in virtual space, accelerating the iteration of high-level planning algorithms that were discussed in Section 4. The programs generated by the low-level controllers, which were discussed in Section 5, can be tested and optimized during the off-line debugging stage. This process significantly reduces the real debugging costs and security risks. In addition, through the construction of digital twins, real-time monitoring, system optimization, and predictive maintenance of the running states of the robots and the entire production line can be achieved.

This section divides the commonly used EIIR simulators into two categories: robot simulators and production-line simulators, and further discusses their limitations and sim-to-real transfer techniques.

## 6.1 Robot simulators

Robot simulators focus on simulating the motion, control, and perceptions of robots [244]. The core function of a robot simulator is to finely model the internal motion mechanism and sensor feedback of a robot. It is based on the integration of a high-fidelity physical engine and robot middleware, such as ROS. It integrates a variety of sensor models and provides high-quality training data for deep learning and reinforcement learning, as well as for other algorithms. Some reviews [2, 6] have discussed robot simulators in relation to the EI field, however, these works primarily focused on service scenarios (such as living room, kitchen, and restaurant scenarios), and the evaluation criteria are uneven, while the industrial field has different requirements. Therefore, this article proposes five primary EIIR-simulator evaluation criteria for industrial scenarios:

1) *High-Fidelity Motion Simulation* (**HFMS**): This indicator is used to evaluate whether the simulator can obtain a simulation effect that is highly consistent with the motion of a real robot. Specifically, it examines whether the simulator is integrated with ROS, which allows the simulator to use control algorithms and motion-planning tools that have been tested by industrial practice in the ROS framework. The use of such tools supports fine and accurate motion simulations of robots of different brands. For software without an official direct ROS interface, this indicator also recognizes that communication between the ROS and the simulator can be achieved through a custom python interface, which compensates for the lack of native support. HFMS not only includes the restoration of the static orientation but also indicates whether the motion responses and mechanical properties that occur under dynamic motion and load changes are consistent with actual robot behavior.

2) *Rich Robot Library* (**RRL**): This indicator primarily focuses on whether the simulator includes many preset robot models. An RRL allows users to directly use preset models for simulation; thus, they do not need to create new robot models or define complex kinematic and dynamic parameters. Industrial applications involve many brands and models of robots. A preset model library can significantly reduce the workload associated with model construction, ensure that many kinds of robots can be quickly verified and used in the simulation platform, improve development efficiency, and ensure the credibility of the simulation results.

3) **Python API**: This indicator is used to determine whether the simulator provides a seamless interface with Python. This is important because Python is widely used for deep learning, reinforcement learning, and data processing. A good Python API allows developers to easily call the simulator functions and to seamlessly integrate the simulation environment with the deep learning training and algorithm-debugging processes.

4) *Multiple Sensor Simulation* (**MSS**): This indicator is used to evaluate whether the simulator is able to incorporate feedback from a variety of simulated sensors that are common in industrial scenarios. Industrial robots often rely upon sensors, such as in-place feedback sensors, proximity sensors, photoelectric sensors, and force sensors, to obtain environmental and state information. High-level MSS not only requires sensors to have sufficient accuracies, but also requires simulations of the response delays, noise characteristics, and interference effects of the sensors. Only in this way can it be ensured that the sensor data in the simulation environment are consistent with the actual application data, which in turn ensures that a real and reliable basis for robot decision-making and action control is provided.

5) **RGB-D**: This indicator determines whether the simulator has a built-in RGB-D camera simulation function. RGB-D sensors can simultaneously collect color images and depth information, thereby providing rich perceptual data for robot vision. Vision systems are important to robot perception, navigation, and manipulation. RGB-D data can be used for object recognition, 3D reconstruction, path planning and environment modeling.

These indicators were used to comprehensively evaluate the performance of robot simulators in industrial applications, as shown in **Table 9**. This evaluation was performed so that the most suitable simulation platforms can be selected for specific industrial scenarios.

**Table 9** Robot Simulator evaluation. In the table, √ indicates that the simulator meets this standard, while ○ indicates that it can be implemented through custom interfaces.

| Environment | Simulator | Year | HFMS | RRL | Python API | MSS | RGB-D | Industrial applications |
|---|---|---|---|---|---|---|---|---|
| Game-based | Gazebo [245] | 2004 | √ | √ | √ | √ | √ | [246-249] |
| | MuJoCo [250] | 2012 | √ | | √ | √ | √ | [251, 252] |
| | CoppeliaSim [253] | 2013 | ○ | √ | √ | √ | √ | [254-256] |
| | PyBullet [257] | 2017 | ○ | | √ | √ | √ | [11, 258-260] |
| | Isaac Gym [261] | 2019 | ○ | | √ | √ | √ | - |
| | Isaac Sim [262] | 2023 | √ | √ | √ | √ | √ | [263] |
| Real-world-based | AI2-THOR [264] | 2017 | ○ | | √ | | √ | - |
| | VirtualHome [265] | 2018 | √ | | √ | | √ | - |
| | VRKitchen [266] | 2019 | | | √ | | √ | - |
| | Habitat [24] | 2019 | ○ | | √ | √ | √ | [123] |
| | iGibson [267, 268] | 2021 | ○ | | √ | | √ | - |
| | TDW [269] | 2021 | ○ | | √ | | √ | - |

Game-based simulators primarily use 3D virtual resources to build the environment, which consists of scenes and objects that are composed of 3D models created in advance. The advantages of such simulators are low resource requirements and rapid scene construction; thus, they are suitable for use in scenarios that do not require a high sense of reality. Particularly in the manufacturing industry, 3D models of various equipment and products are generally obtained during the design stage, and these models can be directly used in the construction of the simulation environment. **Gazebo** [245] is a powerful open-source simulation platform that is closely integrated with the ROS to support high-fidelity motion simulation. It includes an official robot library and provides multi-sensor support; therefore, it is especially suitable for the simulation of multi-robot collaboration in industrial settings. The **Mujoco** [250] is famous for its high-precision physical engine, which is suitable for robot control and reinforcement learning tasks. Its accurate dynamic simulation causes it to be widely used in academia. **Pybullet** [257] is a lightweight physical engine that is suitable for rapid simulation and algorithm testing, particularly for reinforcement learning. Its simple API and Python support reduce the user threshold. The **Isaac Sim** [262] provides high-fidelity motion simulation of multi-sensor systems. It is based on the NVIDIA Omniverse platform, which provides a comprehensive performance. In general, these simulators can provide the necessary functions and support for the simulation of robot motion, perception, and task execution in industrial applications. The differences between them are primarily reflected in the HFMS and RRL indicators. If the application scenario focuses on close integration with the ROS and includes multi-robot collaboration tasks, Gazebo is an ideal choice. However, Isaac Sim has stronger physical-simulation and complex-environment modeling capabilities. **Table 9** presents the primary functions of six game-based simulators. **Fig. 10** depicts some industrial application cases for these game-based simulators.

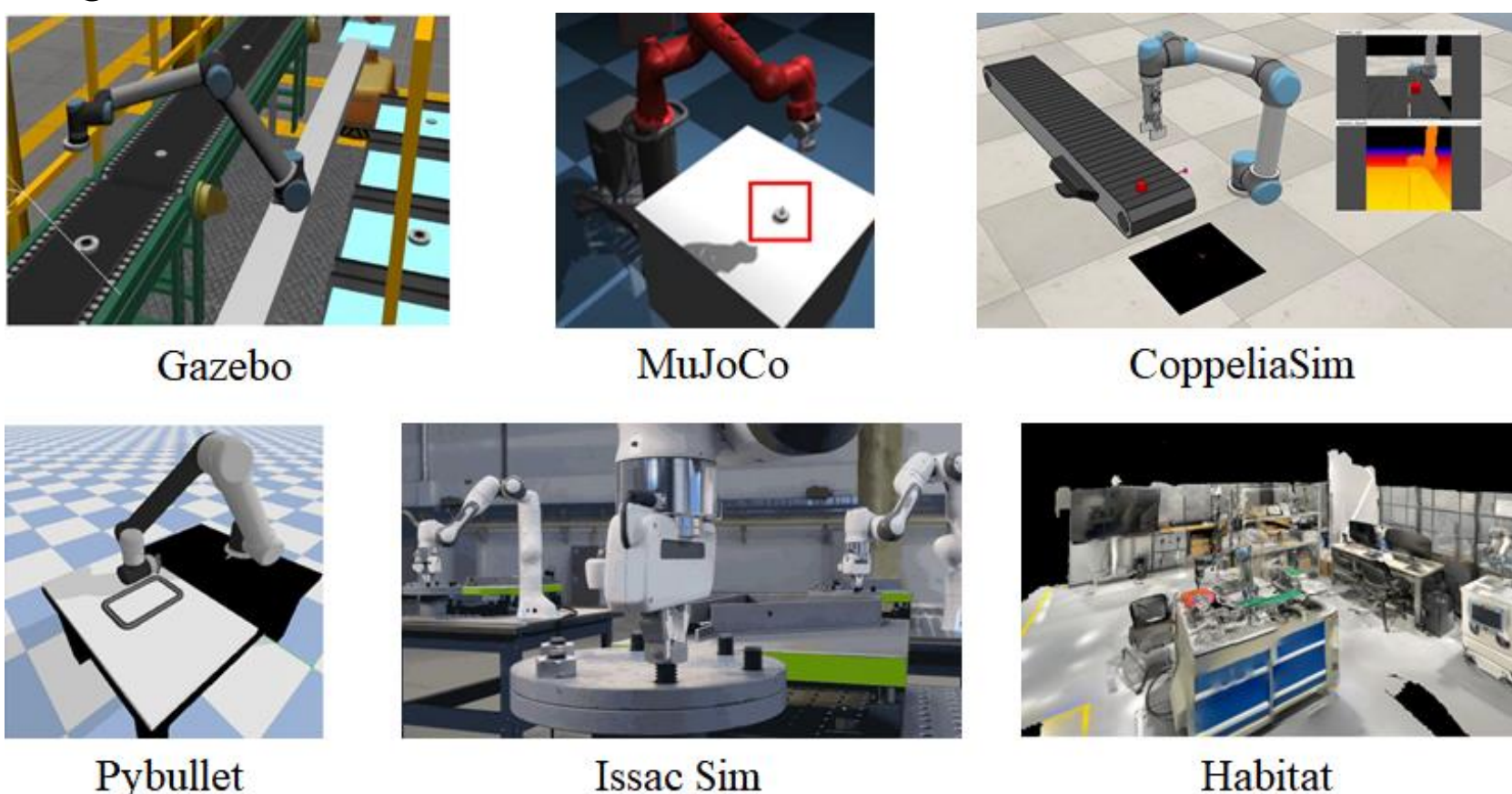

**Fig. 10.** Industrial examples of robot simulators. The figures were obtained from the literature or case studies in **Table 9**.

Real-world-based simulators usually build simulation environments from real-world scanning data. Unlike game-based simulators, they rely on 3D-scanning techniques to transform real-world environments into digital models; thus, they provide a higher sense of reality and greater object detail. Due to their high fidelity, real-world-based simulators are often used in application scenarios that require higher simulation accuracies and Sim2real migration; such applications could include navigation and interaction in indoor environments. **Table 9** presents six real-world-based simulators. The evaluation results show that these kinds of simulators generally use humans as agents during operation, and they do not contain robot libraries. However, they possess many real-world 3D resources, such as furniture, home appliances, and indoor layouts; therefore, they support multi-agent simulation interactions. In addition, real-world-based simulators generally lack the support of sensors that are commonly used in industrial manufacturing scenarios. This lack makes it difficult for these simulators to meet industrial-automation requirements. Of these simulators, only **Habitat** [24] provides a certain degree of EIIR simulation capability; thus, it is the choice with the greatest potential for industrial-robot applications in real-world-based simulators.

In real industrial applications, the simulators described above have been widely used for various types of robot-task simulations, particularly for those associated with automated production environment. For example, **Gazebo** has been used to simulate robot grasping and classification tasks on a conveyor belt. These operations included RGB-D camera-based object recognition and processing. These simulations were performed to verify the performance and task execution efficiency of the robot in a dynamic environment [246, 249]. **Isaac Sim** has been used as the basis of a generative simulation system to provide core support for large-scale manufacturing robot training data [263]. With its powerful physical-simulation capability, this system can simulate various types of robots, manipulation tasks, and manufacturing environments. It can also simulate abnormal robot behavior and abnormal conditions that may occur during manufacturing processes. **Habitat** uses 3D point clouds to reconstruct and annotate human–machine collaborative manufacturing scenarios. It also supports path planning and navigation tasks [123]. Specifically, as an AI simulation environment, Habitat supports the integration of LLMs for the understanding of natural language and the generation of robot actions. The Pathfinder module is also stored in Habitat; this module help AGV to achieve path planning from the current position to the target position.

Although the simulators described above fulfill important functions during industrial-robot task simulation, they still encounter significant challenges when used for production-line tasks. Robot simulators are primarily concerned with simulating the behavior of the robots themselves. However, industrial production-line scenarios involve more than the motion control of the robots. In such scenarios, the relationships between multiple devices must also be integrated and coordinated. For example, in addition to the robot controller, industrial automation usually relies on a PLC for the management of the other equipment. Therefore, to apply EI to actual production lines, the functions of the existing simulation platforms must be further expanded and close integration with industrial automation systems must be considered.

## 6.2 Production-line simulators

Production-line simulators are primarily used to simulate the operation of entire production lines rather than only robots. Unlike robot simulators, production-line simulators focus on coordination between multiple devices, robots, sensors, actuators, and control systems; thus, they are more suitable for industrial scenarios than robot simulators. Their software usually integrates the controllers of various industrial equipment and robots, and they can simulate the workflows of entire production lines. These workflows include various processes, such as material handling and assembly. Since production-line simulators already have their own robot controllers and device interfaces, they do not require integration with the ROS; rather, focus on the simulation of device collaborations in industrial automation scenarios.

Because of these characteristics, the evaluation criteria for production-line simulators are different from those for robot simulators. For example, the HFMS indicator no longer focuses on whether there is integration with the ROS; rather, it determines the number of robot controllers that are supported by the simulator so that the applicable robot-brand range can be evaluated. The RRL indicates the number of robot models that are provided in the simulator or on the official website; thus, it reflects the ability of the simulator to support various robot models. Two additional evaluation metrics, PLC and Multi-devices, are also used to evaluate production-line simulators; they are used to determine whether the simulator can effectively simulate a PLC for production-line control and model the interactions between multiple devices. **Table 10** presents the evaluation results for 10 production-line simulators using the evaluation metrics that were discussed above. The relevant information was gathered from official websites, instructions, relevant research, and case blogs. The relevant values and standards may change due to iterative software version updates, however.

**Table 10** Production-line simulator evaluation. In the table, √ indicates that the simulator meets this standard. For Python API, ◯ indicates that the simulator can be implemented through custom

interfaces. For RGB-D, ○ indicates that the simulator only supports RGB image acquisition.

| Simulator | HFMS | RRL | Python API | MSS | RGB-D | PLC | Multi-devices |
|---|---|---|---|---|---|---|---|
| KUKA.Sim [270] | for KUKA robots | | √ | √ | ○ | √ | √ |
| RobotStudio [271] | for ABB robots | | √ | √ | √ | √ | √ |
| ROBOGUIDE [272] | for FANUC robots | | ○ | √ | √ | √ | √ |
| MotoSim [273] | for Yaskawa robots | | √ | √ | | | √ |
| Robotmaster [274] | 22 | 534 | | | | | √ |
| RoboDK [275] | 40+ | 1295 | √ | | ○ | √ | √ |
| ArtiMinds RPS [276] | 6 | 50+ | | √ | √ | √ | √ |
| DELMIA [277] | 20 | 2000+ | ○ | √ | ○ | √ | √ |
| Visual Components [278] | 17 | 1900+ | √ | √ | ○ | √ | √ |
| Tecnomatix [279] | 18+ | 940 | ○ | √ | √ | √ | √ |

Most of the major industrial robot manufacturers have developed adaptive industrial simulators for their own robot brands. For example, **KUKA.Sim** is a simulation software that is used specifically for the offline programming of KUKA robots [270]. This software can display the robot motion in virtual environments before the equipment is put into operation, which enables motion optimization from the perspective of beat time. It also ensures the feasibility of robot programs and layouts through accessibility checks and collision recognition functions. In addition, because the software supports MSS, PLC, and multi-device interactions, KUKA.Sim can create digital twins; that is, scenes that are exactly like the real production lines. The virtual and real control systems use the same data for their operation. Therefore, it is able to test and optimize new production lines within virtual environments, KUKA.Sim has become the basis of virtual commissioning. The **RobotStudio** software [271], which was developed for ABB robots, and the **ROBOGUIDE** software [272], which was developed for FANUC robots, can accomplish similar functions. However, the **MotoSim** software [273] does not support connections with external PLCs and cannot completely simulate the operation of an entire production line. In addition, EIIR requires a simulator to support robot deep learning. The four software packages described above can directly or indirectly support Python API, which is convenient for integration with external systems or deep learning modules. However, KUKA.Sim and MotoSim do not support RGB-D cameras in the simulation environment; thus, the robots are not able to visually perceive their environment. With respect to production-line simulation and deep-learning support, RobotStudio has the best comprehensive performance.

In addition to the simulators discussed above, each of which is dedicated to a single robot brand, there are a series of production-line simulators that can integrate many robot controllers and post-processors after receiving authorization from various robot manufacturers so that offline programming can be performed. For example, the **Visual Components** simulator integrates 17 post-processors and more than 40 robot controllers, which can be used to control

ABB, KUKA, FANUC, and UR robots, among others. Thus, there is no need to use multiple software types or to understand multiple robot programming languages [278]. In addition, the online model library, eCatalog, of Visual Components contains more than 1,900 robots that can be used in the simulation. Visual Components also supports the integration of sensors and PLC, and it can accurately reflect the real control system used by the physical machine in the model, thereby achieving virtual commissioning of the production line. Visual Components and Python are also closely integrated. The Python script editor can be directly called in the software, and robot control, trigger setting, and signal events can be performed in combination with API. **Tecnomatix** is a similar integrated simulator that serves as the core industrial software of the Siemens Xcelerator digital ecosystem. It is highly compatible with the Siemens PLC, the SCADA system, and the MES/MOM platform [279]. It can accurately analyze robot motion trajectories, beat time, and production-line obstructions, and it is particularly capable of managing complex process workflows, such as multi-robot collaborative welding and flexible-body assembly. In addition to its production-line simulation capability, Tecnomatix also supports RGB-D cameras in its simulation environment. Thus, it can simulate the imaging characteristics of real industrial vision systems by means of a ray-tracing algorithm, and it supports the training and verification of defect-detection algorithms. The virtual reality (VR) system inside the software not only provides an immersive factory-roaming experience but also integrates motion-capture and ergonomic-analysis tools. A series of innovative functions in Tecnomatix are suitable for use by researchers in the further exploration and application of EIIR in industrial settings. **Fig. 11** depicts some industrial application cases of these production-line simulators.

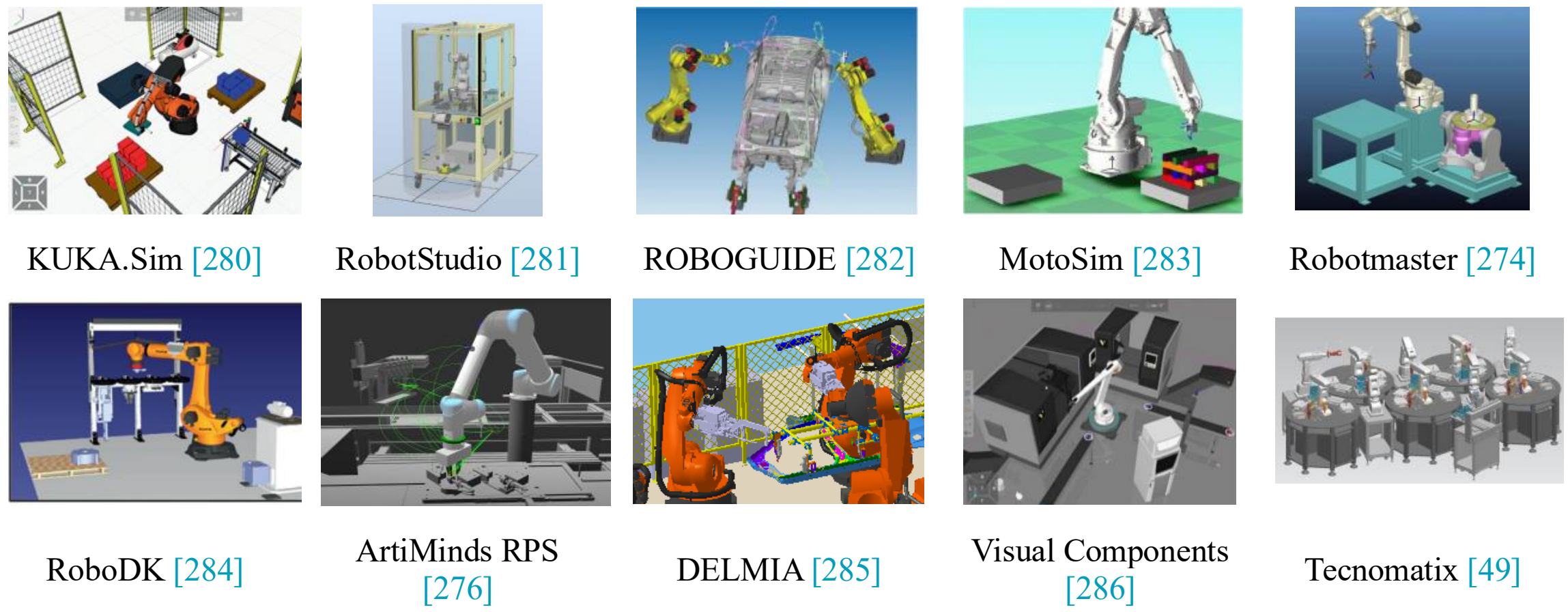

**Fig. 11.** Industrial examples of production line simulator. The figures were obtained from the literature or case studies.

In conclusion, although the existing mainstream production-line simulators tend to be mature with respect to industrial-equipment integration and production-line virtual commissioning, there is still a significant gap between their current states and EI integration. On the one hand, such platforms generally lack open Python API and extensible deep-learning frameworks; thus, it is difficult for them to support the training and testing of decision algorithms. On the other hand, their key modules for real physical interactions, such as deep camera simulation and tactile feedback simulation modules, are not yet perfect; therefore, it is difficult for agents to obtain nearly real sensor input in the virtual environment. Although the robot simulators described in Section 5.1 perform well during single-robot deep-learning training and high-fidelity data acquisition, they are limited by multiple problems, such as their lack of compatibility with PLC communication protocols and the weak linkage capability of the MES. Therefore, they cannot build complete industrial scenarios that include transmission lines, sensor networks, and other elements. For these reasons, when building an EIIR simulator, developers must fully evaluate the differences between the two types of simulators. In cases where current production-line simulators do not natively provide AI-oriented interfaces, production-line simulators can be used to first build a production-line digital twin base with accurate equipment models. After that, developers may mimic the integration practice in real factory environments by using intermediary software (e.g., ROS or OPC servers) to access a dedicated robot simulation node that supports deep learning [287-289], and finally to form a composite simulation architecture that accounts for the fidelity of the industrial equipment and the flexibility of the agent training.

### 6.3 Sim-to-real techniques

The sim-to-real problem refers to the challenge of transferring models or policies trained in simulated environments to the real world, where differences in physical dynamics, sensor noise, and environmental variability can lead to significant performance degradation. Previously, we introduced both robot simulators and production-line simulators, which allow for the efficient collection of large-scale training data in virtual environments. These tools help circumvent the high cost, precision demands, and time-consuming nature of collecting real-world data. However, despite the integration difficulties discussed at the end of Section 6.2, both types of simulators face a common limitation: the difficulty of transferring simulation-trained models to real-world deployment. Because simulation environments inherently differ from the real world, particularly in terms of physics. Most simulators struggle to accurately reproduce complex physical interactions such as friction, material deformation, or noisy sensor feedback. As a result, models trained solely in simulation often fail to generalize to real-world systems, and simulators are typically limited to use as algorithm prototyping or debugging tools.

In recent years, an increasing number of studies have been dedicated to bridging the gap between simulation and real-world deployment. The relevant methods can be broadly categorized into three typical paradigms: tuning simulation, aligning data, and merging policies.

1) *Tuning simulation*: aims to enhance the transferability of policies by adjusting the simulation environment to make the generated simulated data more similar to the real world. The most common approach is domain randomization [290-292], which enhances the robustness of the model by introducing various randomized parameters into the simulation. These may include changes in object categories, quantities, and positions, or variations in textures, lighting, camera angles, reflection properties, and image noise, thereby covering a wider distribution of real-world variations. Another representative method is system identification, which optimizes the dynamics and visual rendering parameters of the simulator using data from real systems. For example, EASI [293] relies solely on a single real demonstration to significantly calibrate the dynamic parameters of the simulator, achieving zero-shot sim-to-real transfer. CAPTURE [294] dynamically adjusts simulation parameters by comparing real interaction data with simulation outputs and employs multiple rounds of iterative optimization to further improve the accuracy of the simulation. In addition, methods such as Real2Sim2Real [295] and VR-Robo [296] are dedicated to constructing high-fidelity digital twins through real-scene scanning and 3D reconstruction, making the simulation environment closer to real-world scenes in terms of geometric structure and visual appearance.

2) *Aligning data*: focuses on processing real data and aligning it with simulated data so that both can be used collaboratively during training. Among them, domain adaptation [297, 298] is the most classic method, which enables the features extracted from simulated and real data to be close to each other in the representation space, thereby narrowing the gap between the training and the deployment distribution. In addition to feature alignment, recent research has also introduced other intermediate representations to enhance the alignment effect. For example, Lang4Sim2Real [299] utilizes natural language as a unified semantic bridge to guide the alignment of visual features between simulated and real images in the semantic space. Specifically, the method pre-trains an image encoder jointly with simulated images and real images annotated with natural language, enabling visual features to have cross-domain semantic consistency. During the policy training stage, it mixes simulated and real demonstration, where real data accounts for only a small proportion, thus achieving few-shot transfer. Another approach utilizes exploration data for virtual-real alignment. For example, the method proposed by Wagenmaker et al. [300] first learns a set of exploration policies covering a wide range of state space through diversified rewards in a simulated environment, and then directly deploys them to real world for exploration data collection. Finally, based on the collected real data, new policies are retrained to achieve indirect transfer from simulation to reality. The common characteristic of above methods is that they do not insist on eliminating the sim-real difference, but rather achieve the connection between virtual and real data by introducing shared intermediate representations, thereby enhancing the practicality of the final policies.

3) *Merging policies*: adopts another approach, neither tuning the simulation environment nor aligning the data distribution between simulation and real-world data. Instead, it trains policies independently in the simulation and real-world environments, and then merges them during the deployment stage. Representative methods such as TRANSIC [301] train a preliminary policy using reinforcement learning in the simulation environment and then directly deploy it to the real-world for execution. During actual operation, when robot behavior deviates or fails, human operators intervene in real-time to correct it, forming a batch of real-world demonstration data with human intervention. Subsequently, the system trains two sub-policies based on simulation data and real-world data respectively, and achieves more stable and smooth control output through a policy merging mechanism. This type of method emphasizes integrated learning at the policy level, rather than unification at the environment or data level, thus providing higher practicality in engineering practice.

In most research on industrial scenarios, researchers are generally committed to building high-fidelity digital twins to achieve effective alignment between the simulation environment and the real factory. Such work essentially belongs to the tuning simulation paradigm in the sim-to-real problem, improving the authenticity and completeness of the simulation environment. Among them, the five-dimension digital twin model [302] is widely used in the manufacturing field as a theoretical framework for building a virtual-real fusion system. This model consists of five dimensions: physical entities, virtual models, data, services, and connections, and can comprehensively describe the structure, state, function, and interaction relationships of industrial systems. Based on the production-line simulator introduced in Section 6.2, a virtual factory with real-time and operable capabilities can be constructed. It not only supports tasks such as virtual commissioning, condition monitoring, and fault prediction but also provides a high-fidelity interactive environment for the training and testing of EIIR. On this basis, combined with other sim-to-real technologies mentioned, the role of the simulation environment in industry can be maximized, significantly reducing the dependence on real data.

# 7 Case study: an EIIR-based flexible assembly system

Based on the EIIR framework we proposed, we aim to deeply integrate EIIR with existing industrial systems, thereby achieving an intelligent and highly flexible system driven by natural language. This section will demonstrate, through a specific case study as shown in **Fig. 12**, how to apply EIIR to the reconfigurable flexible assembly system we previously proposed [49, 241, 303]. In fact, we are promoting the practical implementation of this integration. This system consists of multiple assembly units and is capable of assembling various valves. By adjusting the number and layout of units and adopting quick-change tools and fixtures, the system can achieve hardware reconfiguration. At the same time, with our self-developed assembly program, it also supports reconfiguration of software logic. Leveraging the core modules in the EIIR framework, using natural language as the interactive entry, the world model as the decision-making core, and virtual–real interactions as its verification basis, this system is gradually achieving task-driven, autonomous decision-making, and flexible response. Next, we will specifically introduce the ideal operating mode of this system to help readers more intuitively understand the practical significance of the EIIR framework. Additionally, a video demonstrating the overall operation process of the system is provided in the supplementary material.

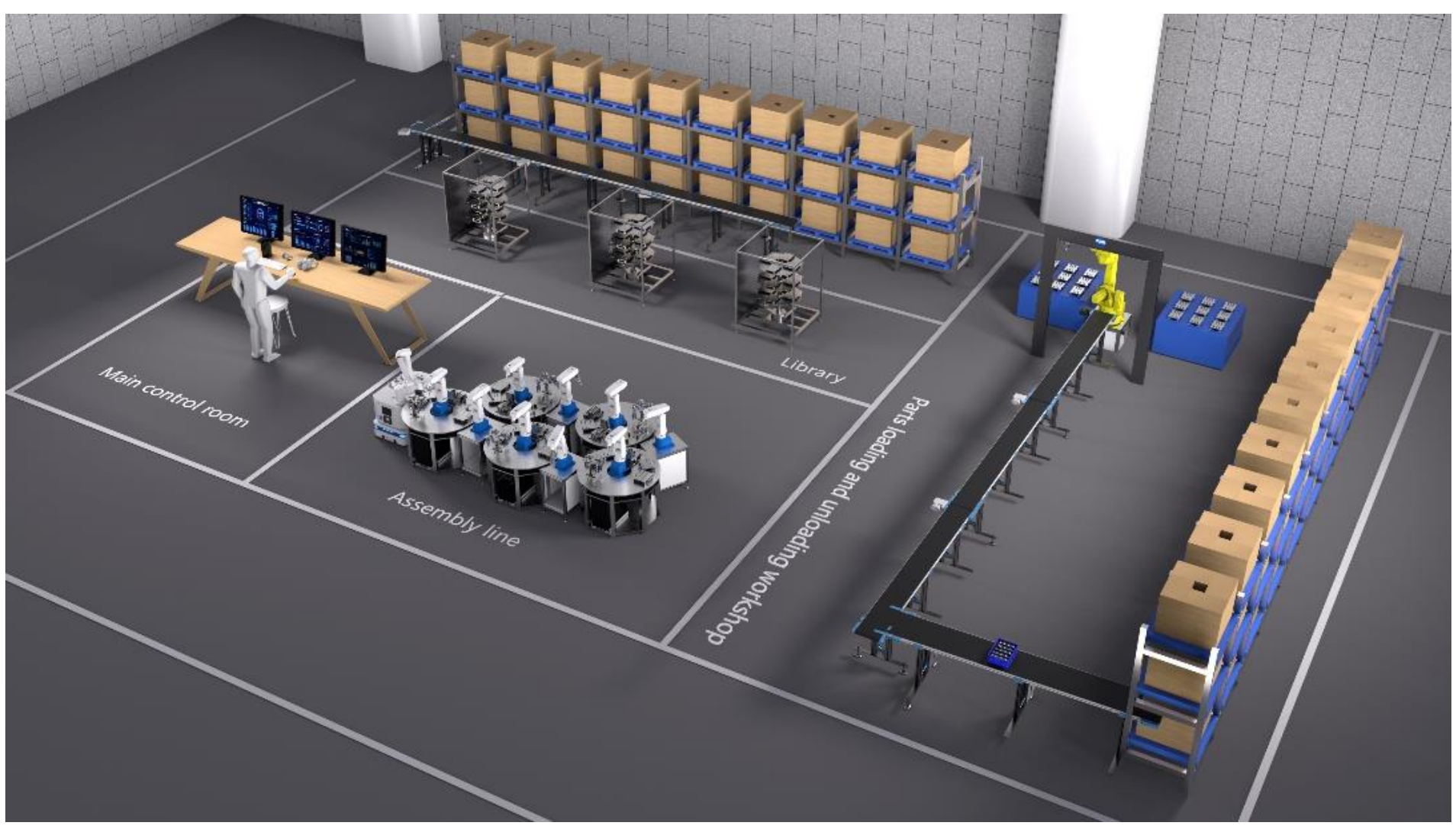

**Fig. 12.** Conceptual validation of an EIIR-based flexible assembly system.

In our envisioned EIIR-based flexible assembly system, the world model serves as the foundation for the system's cognitive capabilities, responsible for comprehending the entire working environment and operating objects. Taking the flexible assembly line as an example, the working area encompasses multiple rooms, such as the part library, tool library, and assembly units. Initially, we automatically construct semantic maps for these areas based on visual information collected by robots [36]. These maps not only contain geometric information necessary for robot navigation (such as walls and passageways), but also embed semantic labels like "Part Library 1" and "Assembly Unit 2". This enables mobile robots to understand what it means to "go to Part Library 1 to retrieve a sleeve and place it on Assembly Unit 2" and from which room and shelf to retrieve the item. Simultaneously, the world model also conducts modeling of the operating-object knowledge for various products. Taking a series of valves as an example, we automatically extract and organize the information originally stored in process documents into a graph-structured assembly knowledge graph. The process is facilitated by our fine-tuned LLM. Besides text, it can also automatically extract geometric parameters as supplementary information in conjunction with CAD models. Ultimately, this information is transformed into knowledge graphs, clearly depicting how and what to use for product assembly, for subsequent reasoning and planning purposes.

With the knowledge provided by the world model, the high-level task planner is to understand the instructions and plan the actions. For example, when the user issues the instruction "Please assemble 1000 pressure-reducing valves.", the large model receiving the instruction will first call upon knowledge embedding, which is the vectorized graph-structured data, to gain a deep understanding of the assembly sequence of valve, the required parts and tools, their locations, and other relevant information. Based on this, it generates a series of reasonable subtasks. These reasoning processes are not carried out in isolation. Our system is equipped with multiple agents that possess equivalent knowledge reasoning capabilities. They propose solutions, then engage in brainstorming-like interactions, evaluation, modification, and optimization, ultimately converging on a complete subtask list that covers all assembly units and material flow. For example, Assembly Unit 1 is responsible for processes 1-3, Assembly Unit 2 is responsible for processes 4-6, and the mobile robot goes to the Part Library to fetch the sleeve and place it on Assembly Unit 1... All of these are clearly defined in the subtask list, taking into account optimization metrics such as production line balance rate, production takt time, and transfer time, laying a solid foundation for the specific execution in the next step.

After generating the subtask list, the next step for the system is to enable the robots to act. The low-level skill controller is responsible for translating these subtasks described in natural language into executable assembly programs. For each assembly unit, each process is mapped to a corresponding program segment based on the rules in the knowledge graph, covering skills invocation, fixture control, tool changing, and other contents. Meanwhile, the skill parameters (such as grasping positions, insertion depths, or torque values) can be automatically populated by combining the topological information stored in the world model with visual observations of the target object. These program segments are then assembled into complete execution programs and issued to the PLCs and robots, enabling rapid adaptation to new product configurations. In addition, we also explore the integration of code-generating LLMs to achieve efficient automatic generation and rapid modification of assembly programs. For specific assembly skills, such as "press-fit a plug sleeve", the system will call the skill model (a VLA model), quickly learn the target skill through several demonstrations and visual inputs, and automatically generate the robot trajectory. For mobile robots executing subtasks such as loading and unloading, we have introduced open vocabulary grasping technology tailored for industrial scenarios, enabling them to understand instructions (such as "take out a sleeve from the Part Library") and identify targets, plan grasping poses, and complete operations in complex stacked environments. At the same time, the robot relies on the semantic map constructed earlier for path planning and autonomous navigation, enabling it to efficiently traverse between different areas and ensure precise and efficient delivery of parts and tools to each assembly unit.

The role of the simulator is to enable the entire system to undergo a virtual rehearsal before actual operation, providing great convenience and safety guarantees for industrial deployment. At the early stage of production line construction, we built a digital twin of the entire assembly system. The digital prototype shares the same appearance, motion mode, sensor configuration, and control logic as the physical system, and can perform the same assembly, loading and unloading tasks. In this way, we can carry out a large amount of work such as task planning verification, path optimization, and program testing without the need for a real machine. In addition, we have also built typical complex industrial scenarios through the simulator to generate training data and carry out algorithm testing. For example, we can simulate the complex stacking layout of parts in the Part Library, automatically generating a large amount of image and pose data to train the robot's embodied grasp strategy [304]. During the operation phase, the simulator can also form a bidirectional digital-twin loop with the physical system: robot data, PLC states, and vision signals are fed back to the simulator for virtual-real synchronization, while virtual sensing and geometric/kinematic models allow the simulator to infer additional production states that are difficult to sense directly from hardware (e.g., the real-time world coordinates of parts). These perceptual signals continuously update the world model, including both operating-object states (e.g., availability, task progress, physical status) and working-environment states (e.g., equipment poses, workspace constraints), ensuring that planning and control are not one-shot open-loop processes. When deviations occur, such as missing parts, misalignment, or tool unavailability, the updated world model can trigger task replanning at the high level or error recovery at the skill level, realizing a perception-decision-action loop that aligns with EI principles.

In summary, this case study demonstrates how the EIIR framework can be implemented in real industrial systems. The entire system is driven by assembly tasks, capable of dynamically generating task execution plans and distributing them to multiple units and robots, achieving flexible collaborative operations in the production line. This architecture design, oriented towards future intelligent manufacturing, not only enhances the system's reconfigurability and adaptability but also provides a practical and feasible technical path and engineering support for application scenarios characterized by "multiple varieties, small batches, and high flexibility" in smart factories. Beyond the presented assembly case, EIIR is also suitable for a wide range of industrial scenarios that involve mult-imodal perception, embodied manipulation, semantic understanding, and dynamic task execution. As illustrated in Fig. 13, our knowledge-driven EIIR framework has already been applied to various real-world tasks in collaboration with our industrial partners. These scenarios span multiple industrial domains, including Computerized Numerical Control (CNC) machining, logistics warehousing, Computer, Communication and Consumer Electronics (3C) electronics, Light Emitting Diode (LED) manufacturing, and new-energy production. The cases encompass three representative categories of embodied tasks in industrial settings: transport, machining, and assembly. This demonstrates that EIIR is not limited to assembly lines but has broad applicability across the industrial chain.

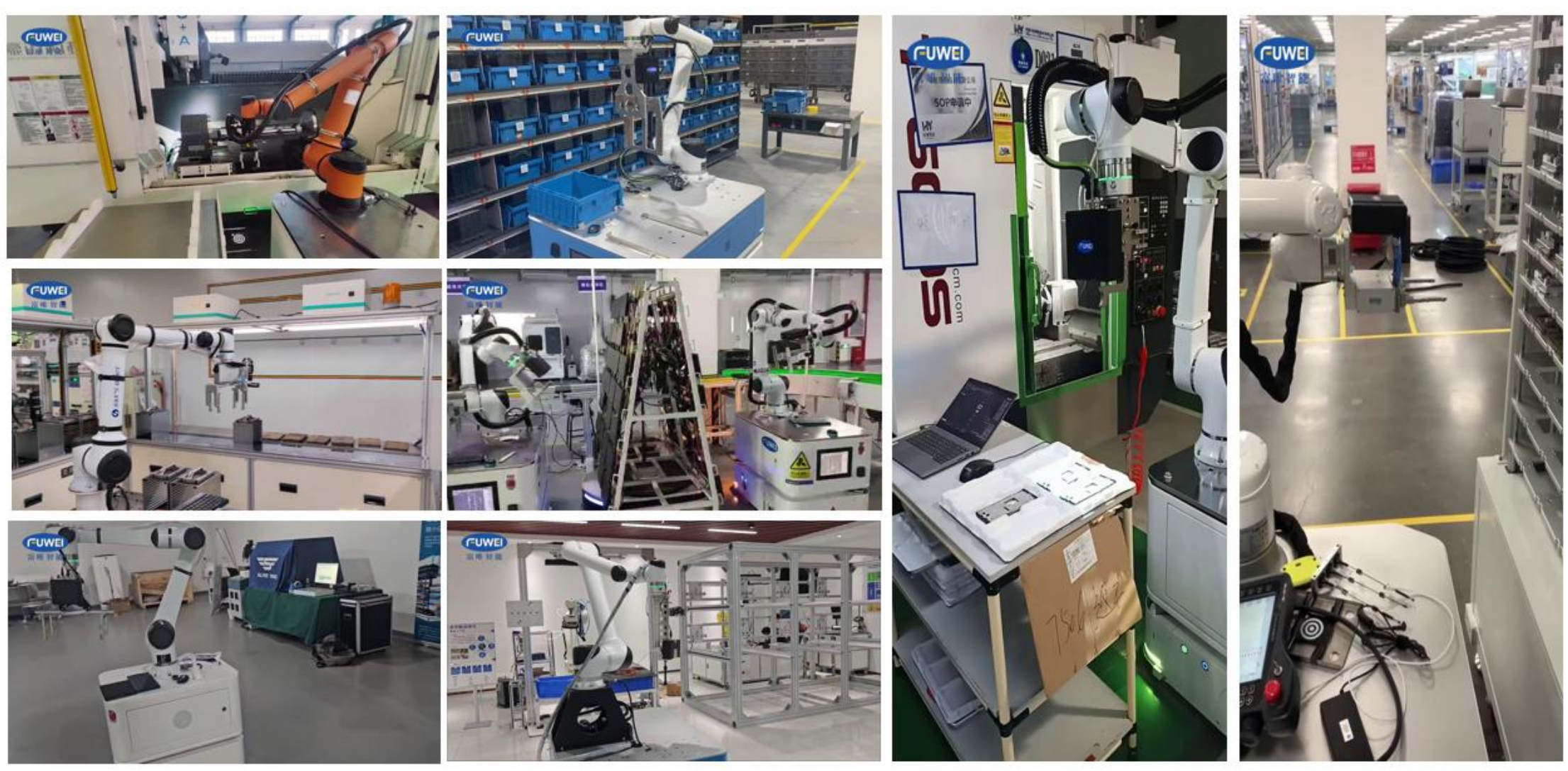

**Fig. 13.** Industrial scenarios suitable for applying EIIR, including CNC machine tending, logistics parcel transferring, 3C component kitting, LED material loading, medical package handling, and screw-tightening in new-energy manufacturing.

# 8 Challenges and future work

According to the development status and trends of the methodologies related to the five modules of the EIIR framework, which were summarized in Sections 3–7 of this paper, the possible challenges and potential directions of future research regarding the application of EIIR techniques to industrial scenarios or systems can be summarized. These summaries are presented in the following sub-sections.

## 8.1 Industrial world model

The authors of this paper believe that, if EIIR is to be successfully deployed in real industrial scenarios, it should possess the three most basic kinds of knowledge: general knowledge, working-environment knowledge, and operating-object knowledge. The existing LLMs are prone to "industrial illusion" when they are used for industrial tasks; that is, the results appear to be semantically correct, but they cannot be used in industrial scenarios. Therefore, the development of an industrial foundation model that can quickly and accurately manage the tasks that are related to the entire life cycle of an industrial scene [305] (which includes product design, manufacturing, testing, and maintenance) is urgent.

With respect to potential solutions, due to the complexity of industrial scenarios and processes, the authors believe that the industrial foundation model should first be decomposed into a series of domain foundation models, such as foundation models for assembly, processing, and product design. Then, a mechanism, such as a mixture of experts (MOE), must be designed to integrate various domain foundation models and thereby form the final output. It is expected that, with the development of an industrial foundation model, the capabilities of LLMs to complete various tasks should reach or even exceed those of experts in various industrial fields. During the training process, the industrial foundation model will also have a large amount of working-environment knowledge and operating-object knowledge from the industrial scene, and this knowledge can also be used to enable the construction of semantic maps and knowledge graphs.

## 8.2 Industrial high-level task planner

In industrial scenarios, the lack of working-environment knowledge and operating-object knowledge has become the core obstruction that restricts EI task planning. The traditional framework relies upon the general knowledge of a large model (an LLM or a VLA model), which can interpret natural language tasks and perceive the positions of objects. However, due to the lack of in-depth industrial knowledge (such as ISO standards and process manuals), the task-decomposition results deviate from engineering constraints (the sequencing of assembly steps may be incorrect or the process requirements may be ignored). The rule-based systems of existing industrial methods are limited because the rigid logic must be defined manually; thus, adaptation to the requirements of flexible manufacturing is difficult. Learning-based methods rely on massive amounts of annotation data, so they cannot be quickly migrated to new production lines.

Therefore, studying RAG-like high-level task-planning techniques that are based on semantic maps and domain knowledge graphs is urgent. Semantic maps and knowledge graphs can structurally store various types of knowledge, such as environmental information, part parameters, and assembly processes, and they can constrain the reasoning paths of the large models. RAG dynamically enhances the domain cognition of the large models through real-time retrieval of external knowledge bases, such as process documents and quality inspection standards. This integration path is expected to overcome the "knowledge blind spot" of existing frameworks and achieve a transformation from general semantic understanding to industrial deterministic planning.

## 8.3 Industrial low-level skill controller

First, the EIIR skills that are capable of generalizing industrial data, such as industrial object detection for open vocabulary, must be studied. To reduce the threshold for industrial customers to use EIIR techniques, industrial data-sensing techniques with stronger generalization capabilities are urgently needed. Taking “6D pose estimation” (which is a “Measure” skill) as an example, most industrial parts consist of a series of parametric parts; that is, these parts are essentially instantiated from a parametric template by using various parameter values, while the primitives of the parametric template and the constraint relationships between the primitives remain constant. However, the existing point-cloud deep-learning methods have not found this data feature. If this method is directly applied to industrial data, its performance declines significantly. Therefore, it is very necessary to thoroughly study perception techniques for 2D [306] and 3D industrial data [307, 308]. Similar requirements for industrial scenarios also exist for other skills, such as the “Pick and place” and “Transport” skills.

Second, it is also necessary to study the general low-level control language for industrial heterogeneous devices. In industrial settings, the EI "body" not only possesses single robots, but it must also cooperate with other devices that are driven by the PLC. However, the existing low-level controller in the EIR framework is limited to the ROS, and the generated standardized action instructions cannot be directly adapted to an industrial controller; therefore, a "protocol wall" is generated for multi-device cooperation. For example, a “grasp” command to a robot must trigger a cylinder clamping action, which is synchronously controlled by the PLC; however, a timing discrepancy between the ROS and the industrial fieldbus can easily cause misaligned actions or safety risks. To eliminate this discrepancy, it is necessary to build an industrial DSL that can serve as an intermediate link between the agents and the physical devices. The primary design objectives of DSL include: Protocol independence, dynamic compilation of the instructions into the native control language of the target device (such as URScript for the robot and ST for the PLC), which enables the seamless connection of cross-brand and cross-type devices; Scalability, support protocol plug-ins that are based on modular architecture, adaptability to the rapid reconfiguration requirements of flexible production lines.

In addition, the skill also faces challenges in fusion and cross-embodiment transfer. In complex and dynamic environments, a single predefined skill is often insufficient to respond to unexpected situations. Therefore, future systems should support reactive skill composition by selecting, adapting, and fusing multiple atomic skills from the library to handle unseen task combinations. Moreover, thanks to the development of high-fidelity simulators with rich robot and sensor libraries, it becomes possible to explore skill transfer across different embodiments, such as transferring learned manipulation skills from a robotic arm to a mobile robot, or even to a humanoid robot. This requires the alignment of perceptual features and control spaces across platforms, and represents a promising research direction for enabling generalized, transferable intelligence in industrial settings.

## 8.4 Industrial production-line simulator

The existing robot simulators (such as Gazebo and Isaac Sim) focus on dynamic modeling and motion simulation for a single robot, while system-level simulation that involves mechanical, electrical, hydraulic, and control multi-domain coupling must be achieved for industrial production lines. It is difficult for the exiting simulators to model such cross-domain interaction processes, and this difficulty results in significant deviations between the virtual commissioning results and actual production-line conditions. In addition, although traditional industrial production-line simulators (such as Tecnomatix and DELMIA) can build high-fidelity production-line digital twins, they cannot support the online training and strategy optimization of embodied agents due to the lack of open deep-learning interfaces.

To meet the dual needs of multi-domain coupling and agent training data, it is necessary to build an EIIR simulator that fuses the virtual and real domains. This process includes two primary directions. First, an open-source simulator should be proposed to support the operation of virtual industrial agents. This next-generation simulator must overcome the limitations of single-robot modeling, build an open platform to support single-robot and production-line agents in industrial scenarios, and support rapid agent deployment. Through open-source community collaboration, the simulator will be able to integrate a variety of industrial control protocols, and it will be compatible with the ROS and deep-learning frameworks, thereby providing a plug-and-play training environment for industrial agents. Second, a simulation data engine for industrial foundation-model adaptation must be established. According to the training requirements of large models, the simulator must strengthen the data-generation capability and generate millions of diversified working-condition data in batches through parametric scene configuration. For example, it must generate robot-trajectory, visual, tactile, and force-sensing data. According to the high-fidelity multi-modal training data discussed above, a closed-loop path, "Equipment-level physical fidelity → Production line-level logic verification → Agent strategy optimization", can be formed.

## 9 Conclusions

By the integration of multi-modal perception, autonomous decision-making, and physical interaction capabilities, embodied intelligent industrial robotics is reshaping the technical paradigm of traditional industrial automation. Rather than relying on rigid manual teaching controls and predefined programs, it now enables flexible, autonomous production. This paper systematically proposes and analyzes the knowledge-driven EIIR framework and summarizes its five modules: a world model, a high-level task planner, a low-level skill controller, a simulator, and a physical system. Comparisons and summaries are provided for the existing working-environment and operating-object knowledge representation methods, the general task-planning and industrial task-planning methods, the skills, the control languages, the simulators, and the sim-to-real techniques. They provide clear illustrations of the latest EIIR developments. To demonstrate the practical value of EIIR, we present a case study on a flexible assembly system. This example illustrates how EIIR can be effectively deployed in real-world manufacturing systems and provides a reference for future industrial applications. The ultimate goal of industrial embodied intelligence is to become the "cognitive center" of intelligent factories, promoting the transition of the manufacturing industry from "program solidification" to "independent evolution." This process requires not only collaborative algorithm and hardware innovations, but also a reconstruction of the industrial software ecology and ultimately the creation of a new industrial paradigm that utilizes embodied intelligence.

## Acknowledgement

This work was supported by National Natural Science Foundation of China (92467204, 61972220), Natural Science Foundation of Guangdong (2022A1515011234), and Shenzhen Major Undertaking Plan (CJGJZD20240729141702003).

## Declaration of competing interest

The authors declare that they have no known competing financial interests or personal relationships that could have appeared to influence the work reported in this paper.

## CRediT authorship contribution statement

Conceptualization and framework, Chaoran Zhang and Long Zeng; investigation, Chaoran Zhang, Chenhao Zhang, Zhaobo Xu, Qinghongbing Xie; writing-original draft, Chaoran Zhang, Chenhao Zhang; case design, Jinliang Hou; writing-review & editing, Zhaobo Xu, Long Zeng; project administration, Pingfa Feng. All authors have read and agreed to the published version of the manuscript.